\documentclass[10pt,journal,compsoc]{IEEEtran}
\usepackage{times}
\usepackage{helvet}
\usepackage{courier}
\usepackage{graphicx}
\usepackage{amsmath,amssymb,amsfonts}
\usepackage{algorithmic}
\usepackage[ruled,linesnumbered]{algorithm2e}
\usepackage{subfigure}
\usepackage{hyperref}
\usepackage{amssymb}
\newtheorem{definition}{Definition}
\newtheorem{problem}{Problem}
\usepackage{color}
\usepackage{gensymb}
\usepackage{stfloats}

\begin{document}

\title{\huge Cooperative Path Planning with Asynchronous Multiagent Reinforcement Learning}
\author{Jiaming~Yin,
        Weixiong~Rao,
        Yu Xiao,
        Keshuang Tang
\IEEEcompsocitemizethanks{\IEEEcompsocthanksitem J. Yin, W. Rao, and K. Tang are with Tongji University, Shanghai, China.\protect\\
E-mail: wxrao@tongji.edu.cn
\IEEEcompsocthanksitem Y. Xiao is with Aalto University, Espoo, Finland.\protect\\
E-mail: yu.xiao@aalto.fi
}
}


\maketitle
\begin{abstract}
In this paper, we study the shortest path problem (SPP) with multiple source-destination pairs (MSD), namely MSD-SPP, to minimize average travel time of all shortest paths. The inherent traffic capacity limits within a road network contributes to the competition among vehicles. Multi-agent reinforcement learning (MARL) model cannot offer effective and efficient path planning cooperation due to the asynchronous decision making setting in MSD-SPP, where vehicles (a.k.a agents) cannot simultaneously complete routing actions in the previous time step. To tackle the efficiency issue, we propose to divide an entire road network into multiple sub-graphs and subsequently execute a two-stage process of inter-region and intra-region route planning. To address the asynchronous issue, in the proposed asyn-MARL framework, we first design a global state, which exploits a low-dimensional vector to implicitly represent the joint observations and actions of multi-agents. Then we develop a novel trajectory collection mechanism to decrease the redundancy in training trajectories. Additionally, we design a novel actor network to facilitate the cooperation among vehicles towards the same or close destinations and a reachability graph aimed at preventing infinite loops in routing paths. On both synthetic and real road networks,  our evaluation result demonstrates that our approach outperforms state-of-the-art planning approaches.
\end{abstract}

\begin{IEEEkeywords}Shortest Path, Route Planning, Multi-agent Reinforcement Learning.
\end{IEEEkeywords}
\maketitle                                                                                          

\section{Introduction}
The classic shortest path problem (SPP) aims to determine an optimal route in a road network for a {given} source-destination pair typically with the goal to minimize {either} travel time or path distance. In a real world, it is common to plan the shortest paths for multiple source-destination pairs (MSD), namely MSD-SPP. 
The challenge of solving MSD-SPP is that we have to consider the traffic capacity constraints of road networks. When the number of vehicles on a certain road segment exceeds the road capacity constraint, traffic congestion occurs, leading to increased travel time. Given the capacity constraints of road networks, the goal of MSD-SPP is to plan the shortest paths for all source-destination pairs with the optimization objective of minimizing an aggregate metric, such as the average travel time {across all paths.

Though the classic SPP problem has been well studied in the literature, solving MSD-SPP is rather hard. Even a simplified {variant} of the problem, e.g., the $k$-Disjoint Shortest Path problem ($k$DSP), when considering only two source-destination pairs, has been proved to be NP-complete  \cite{disjoint-spp}. Here, the $k$DSP aims to find disjoint shortest paths for $k$ source-destination pairs to alleviate congestion on graph edges. In real-world scenarios {where $k$ is large}, {solving} $k$DSP becomes much harder. It is particularly true {since} road segments {may need to accommodate} multiple vehicles {simultaneously while subjecting} to capacity constraints.

\begin{figure}[htbp]
\centering\vspace{-2ex}   
\includegraphics[width=1.0\linewidth]{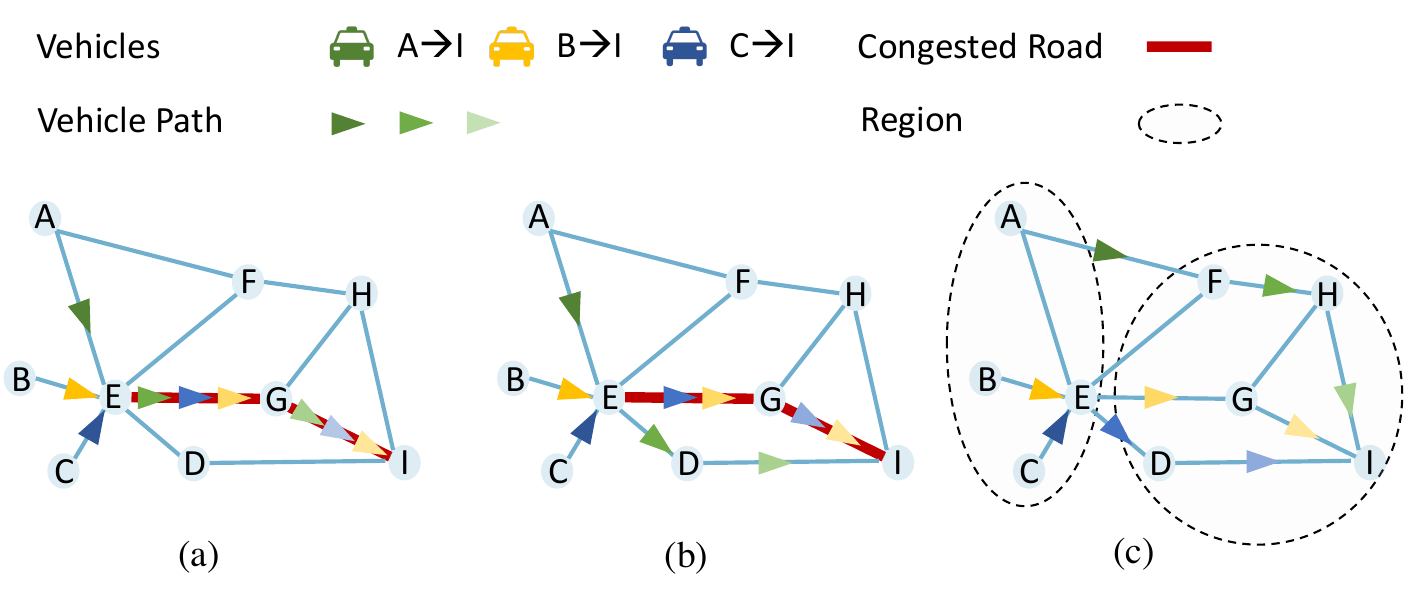}
\caption{Illustrative comparison among (a) Entire Path-based methods, (b) Next-hop node-based methods, and (c) our method. Each node represents an intersection and each edge represents a road segment.}
\label{fig:msd_example}
\centering
\end{figure}


To illustrate the challenges {inherent in} the MSD-SPP problem, we {present} two baseline approaches: (1) The \textit{Entire Path-based Approach} independently plans {the} entire routing path for {each} individual vehicle. For example, we can exploit a classic algorithm such as Dijkstra's\cite{Dijkstra59} algorithm and A$^*$ search algorithm \cite{Astar} to compute the shortest path for {a given} source-destination pair. However, when considering  multiple source-destination pairs in {the context of} MSD-SPP, this approach {may result in the planning of shortest paths that share} certain road segments. When the number of vehicles {on these} road segments exceeds the capacity constraint, traffic congestion occurs. In Figure \ref{fig:msd_example}(a), the road segments $E\to G$ and $G \to I$ become congested because all three vehicles are {routed to these segments simultaneously}. 

 (2) The \textit{Next-hop node-based approach} re-plans routes for individual vehicles upon reaching intersections to select the next intersection based on the real-time traffic conditions. Some deep reinforcement learning (DRL) algorithms such as \cite{darp, yin2023learn} have been proposed to learn such routing policies, which enables adaptive decision-making in response to changing traffic conditions.
 As illustrated in Figure \ref{fig:msd_example}(b), when the green car arrives at the intersection node $E$, we can exploit a DRL algorithm to choose the intersection $D$ as the next intersection, when observing congestion on the road segment $E\to G$. However, it still does not effectively handle the situations where multiple vehicles simultaneously enter an intersection. Again in Figure \ref{fig:msd_example}(b), when the yellow and blue cars arrive at the node $E$ simultaneously, the DRL model might select intersection G as the next hop for both cars, thereby potentially exacerbating congestion on the road segment $ E \to G$. Another limitation of next-hop node-based approaches is that they may generate infinite loops, due to the lack of a global view.

Some recent studies exploit multi-agent reinforcement learning (MARL) to enable the cooperation among multiple agents by treating either vehicles or intersections as agents \cite{xrouting, sigspatial22}. Such cooperation offers promising results in planning vehicle routes. However, these works still suffer from the two issues. 

\begin{itemize}
\item The MSD-SPP problem is an \emph{asynchronous} decision-making setting, where vehicles may not arrive at intersections simultaneously.  In Figure \ref{fig:msd_example}(b), when the yellow and blue cars have already arrived at the intersection $E$, they require the agent at node $E$ to decide the next hop for each of them. Yet, the green car is still on its way to $E$, and no decision is required until its arrival at $E$. However, the centralized training and distributed execution (CTDE) framework \cite{cooperativeMARL} widely used in the MARL literature assumes that the agents synchronously make actions and collects joint observations and actions of all agents at each step. If assuming synchronous decision-making among agents at each time step, it is notable that each vehicle requires multiple time steps to complete an action and reach the next intersection. As vehicles are still its way to the next intersection, the action information gathered at each time step remains unchanged. This leads to redundant data within the training trajectory data, making the training process inefficient. As a result, the CTDE framework does not work well in the asynchronous MSD-SPP setting.
\item When the number of intersections or vehicles is large, existing works using intersections or vehicles as multi-agents encounter \emph{scalability} issues and the MARL algorithms struggle to perform effectively on real-world road networks with thousands of intersections and vehicles.
\end{itemize}

Until now, the literature works above suffer from the ineffectiveness caused by poor cooperation and the scalability issue. To tackle the issues above, in this paper, we propose an effective and efficient framework to solve MSD-SPP. 
Firstly, in terms of scalability issue, we propose a two-stage route planning framework. Specifically, we divide the graph representing the original road network into sub-graphs, also known as regions. By modeling each region as an agent, we exploit MARL to develop a region-level route planning approach. That is, for every SPP task within MSD-SPP starting from a given region, the route planning consists of two stages: 1) the \emph{inter-region route planning}: the agent of the starting region chooses a cutting edge to a neighbour region, and 2) the \emph{intra-region route planning} plans the sub-path from the source node towards the entry point of the selected cutting edge on the sub-graph corresponding to the region. We repeat the route planning steps above until the arrival at the final destination region. In this way, we decompose the original MSD-SPP into the calculation of multiple sub-paths on small sub-graphs and can scale to large graphs with hundreds or thousands of nodes. The key of our framework is to decide the start nodes and end nodes of the sub-paths, i.e., the inter-region route-planning.

Given the two-stage route planning framework, we develop a novel asynchronous MARL framework, namely asyn-MARL, to make cooperative inter-region route planning. The asyn-MARL model builds three following components. \emph{1}) To address the asynchronous issue and maintain the stability of MARL training, we extend the classic CTDE approach to the asynchronous settings. We first design a global state, which exploits low-dimensional hidden vector to implicitly represent the high-dimensional features of joint observations and actions of multi-agents during the MARL training phase. 
Additionally, to decrease the redundancy in the MARL training trajectories, we develop an asynchronous MARL training trajectory collection mechanism for more efficient policies. 
\emph{2}) To enable the cooperation of vehicles and avoid the plan of too many vehicles on the same road segments, we employ a novel actor network, where a GRU module is applied to the routing requests from multiple vehicles  received by the agent at the same time, to extract the features about the competitions among such vehicles. \emph{3}) To mitigate infinite loops during the cooperative route planning, we design a reachability graph to prune actions that may lead to loops. 
}As summary, we make the following contributions.

\begin{itemize}
\item We propose a scalable route planning framework, consisting of two stages of inter-region and intra-region planning schemes. By extending the fine grained next-hop node-based planning model, the region-level route planning framework can work efficiently on large road networks. 
\item We develop an asyn-MARL framework to make cooperative route planning in the asynchronous settings. The asyn-MARL framework consists of an carefully designed global state to represent the joint observations and actions, an asynchronous trajectory collection mechanism to decrease redundant trajectories, and a novel actor network and a reachability graph to mitigate traffic congestion. To the best of our knowledge, this is the first asynchronous MARL route planning model to solve MSD-SPP.

\item We conduct experiments on both synthetic and real-world traffic networks with a microscopic traffic simulation environment SUMO. Evaluation results demonstrate that our approach outperforms both traditional SPP algorithms and state-of-the-art MARL approaches in terms of throughput, travel time and CO2 emission volume.
\end{itemize}

The rest of this paper is structured as follows. Section \ref{sec2:relate} reviews preliminaries and related works. Following that, Section \ref{sec3:def} outlines the problem definition and presents the overall framework. Section \ref{sec4:detail} delves into the details of our proposed solution. In Section \ref{sec5:experiment}, we evaluate the effectiveness of our approach, and finally, Section \ref{sec6:conclude} concludes the paper.

\section{Related Works}\label{sec2:relate}

{This section begins with} a preliminary overview of cooperative MARL in Section \ref{sec22:comarl}, followed by a review of the state-of-the-art for vehicle shortest path planning in Section \ref{sec21:spp}.

\subsection{Overview of Cooperative MARL}\label{sec22:comarl}

Numerous MARL algorithms have been designed to address the scenarios that involve interaction among multiple agents. A straightforward approach is to train each agent independently to maximize their individual rewards by treating other agents as part of the environment \cite{idqn}. However,in traditional RL, the environment is typically assumed to be stationary, meaning that the dynamics of the environment remain constant over time. However, in MARL, each agent's actions affect the environment, which in turn influences the behavior of other agents. This interaction between agents introduces \emph{non-stationarity} into the environment, as the environment changes caused by the actions of all involving agents. Due to the non-stationarity of the environment, the Markov assumption that the current state of the environment contains all the information necessary to make decisions about future actions, does not hold in the context of cooperative MARL.\cite{nonstation19}.

To address the \emph{non-stationarity} issue in MARL, cooperative MARL systems \cite{cooperativeMARL}, such as centralized critic \cite{coma, maddpg, mappo} and value function factorization \cite{vdn, qmix, qtran}, have been developed. A centralized critic is a mechanism where agents share a common critic or value function that estimates the expected cumulative reward for each agent, taking into account the observations and actions of all agents.
For example, Lowe et al. \cite{maddpg} proposes a multi-agent policy gradient algorithm where agents learn a Deep Deterministic Policy Gradient (DDPG) policy \cite{ddpg}. This approach ensures a stationary environment even if the policies of other agents change by using a centralized critic with the joint observations and actions of all agents as input. Next, Yu et al. \cite{mappo} demonstrates that the on-policy RL algorithm, i.e., Proximal Policy Optimization (PPO) \cite{ppo}, performs well in cooperative multi-agent settings with a centralized critic.
Some value-based methods for MARL such as VDN\cite{vdn} and QMIX\cite{qmix} train decentralised policies in a centralised fashion by employing a network that estimates joint action-values as a combination of per-agent values that condition only on local observations.

In the cooperative MARL systems above, a commonly used approach is the so-called centralized training decentralized execution (CTDE) paradigm. It often takes the joint actions and observations of all agents as input to the centralized critic. The CTDE paradigm works wells in the setting of Decentralized Partial Observable Markov Process (Dec-POMDP) \cite{DECPOMDP}. It assumes that all agents perform actions at each time step, indicating the actions of such agents are all \emph{synchronous}. Yet, in typical vehicle routing tasks, this assumption does not hold and instead agents frequently make decisions \emph{asynchronously}. Due to the redundancy of joint observations and actions within the training trajectory, several observations may correspond to the same action, or conversely, a single observation may correspond to multiple highly diverse actions. As a result, the CTDE paradigm does not work well in the asynchronous setting.

To address this issue, some asynchronous MARL methods are developed. Wang and Sun \cite{ijcai-21} 
develops an asynchronous MARL model to solve the bus fleet control problem. It designs a critic to effectively estimate the contribution of the other agents in the asynchronous setting. \cite{aamas23} considers the cooperative exploration of robots. To address the issue that robots accomplish actions at different time steps, this work extends multi-agent PPO to the asynchronous setting and applies action-delay randomization to improve the generalizability of learned policies to action delays in the real world. In the cooperative e-vehicle charging problem, \cite{asmppo} proposes to collect the transition trajectory of an agent based on its own local clock and uses the concatenation of the locations and remaining charging time of other agents as the state.  Nevertheless, these works do not perform well in our asynchronous MSD-SPP setting. 
Firstly, the works are essentially scenario-specific and are inapplicable to our MSP-SPP problem due to the significantly different observation space.
Secondly, unlike these previous works, the asynchronous MSD-SPP problem suffers from two following issues: 1) the number of vehicles arriving at intersections or regions differs depending upon time steps and associated agents, and 2) the number or equally dimensionality of decision actions also dynamically differs depending upon time steps and associated agents. How to address such issues is non-trivial.

\subsection{Vehicle Shortest Path Planning}\label{sec21:spp}
Existing works on shortest path planning {for vehicle navigation} typically aim to plan shortest paths and meanwhile avoid traffic congestion. Depending upon whether or not the route planning cooperates among vehicles, we divide these works into the following categories.

\textit{Non-cooperative route planning approaches} compute the SPP for an individual vehicle with no consideration of concurrent planning for other vehicles. The classic approaches, Dijkstra algorithm \cite{Dijkstra59} and A* search algorithm \cite{Astar}, compute the shortest path between a source-destination pair on a road network. Recently, to solve the classic combinatorial optimization problem
such as Travel Salesman Problem (TSP) and Vehicle Routing Problem (VRP), the machine learning-based approaches learn an approximation function that maps input road networks to output travel tours \cite{ptrnet,vrp_kool}. Nevertheless, they do not work well on dynamic road networks. That is, whenever either graph typologies or edge weights of road networks change, these algorithms have to re-plan the shortest paths, leading to  high computing overhead. 

When a vehicle arrives at an intersection node, DRL-based approaches instead iteratively re-plan the {next hop} based on current traffic states. As a result, the number of re-plans is just equal to the intermediate nodes within the path between the source-destination pair. For instance, our previous work \cite{darp} employs a dueling deep Q-network to determine the shortest path-based vehicle routing on a grid road network. The work \cite{SPPGS} exploits a graph convolution network and a deep Q-network to perform the shortest path-based routing on dynamic graphs. Unlike the simple SPP problem, our previous work \cite{yin2023learn} studies the NP-hard constrained shortest path problem on dynamic graphs. 

All the works above \emph{independently} compute or plan the shortest paths for an individual vehicle with no cooperation {with} other vehicles. Such paths could lead to traffic congestion on certain road segments if an excessive number of vehicles are unfortunately planned onto the same road segments.

\textit{Cooperative route planning approaches} consider the cooperation among vehicle route planning. The $k$DSP problem \cite{disjoint-spp}, with the aim to find disjoint shortest paths for $k$ source-destination pairs, is NP-complete even with only $k=2$ source-destination pairs. 
The classic Gawron algorithms \cite{gawron1998iterative, iterative_mix} find an approximation solution to the optimum. Given the traffic demand between intersections, in each iteration, these methods compute the fastest route for each vehicle and then assign a cost to each road segment based on the intensity of traffic. By iteratively moving some traffic to less congested paths and re-computing road costs, they have chance to finally achieve a user equilibrium. However, these methods are computationally expensive due to the iterative steps.
Instead of computing all routing paths directly, \cite{learn_to_route} and \cite{gddr} develop a RL policy to assign edge weights, and then exploit a softmin function to convert edge weights into flow ratios on graph edges. The two works greatly reduce the solution space from 
$V(V-1)E$ to $E$ where $V$ and $E$ are the numbers of vertices and edges, respectively. Nonetheless, for a large graph size, the training of reinforcement learning policy networks is still hard due to large action space, suffering from poor scalability.


Instead of computing all shortest paths together, MARL-based methods attempt to learn decentralized routing policies with multiple vehicle or intersection-based agents. Since the action space of a single agent is much smaller than the original solution space, it is feasible for MARL to train such policies efficiently.

Depending on how agents are modelled in the MARL setting, these methods can be split into two following categories. (1) Some works assign agents to individual \emph{vehicles}. By treating each vehicle as an agent, \cite{shou2022multi} develops a mean field multi-agent deep Q learning (DQN) algorithm to update its en-route path choice when the multi-agents interact with each other in road networks. On autonomous vehicles, \cite{xrouting} learns a routing policy by incorporating a transformer network \cite{transformer} into the actor model to select the next-hop road segment. The previous work \cite{icpads19} proposes a stage learning algorithm to learn the $\gamma$-Nash equilibrium in the MSD-SPP setting. However, these methods work well with a small number of vehicle agents, and only deploy multi-agents on a small amount of vehicles (typically in tens or hundreds). (2) Some works instead treat each \emph{intersection} as an agent and provide routing decisions for incoming vehicles. For instance, \cite{sigspatial22} assigns an agent to each road intersection and selects a next-hop road segment when a vehicle arrives at the intersection. The work \cite{navtl} makes cooperative control decision for both traffic signal lights and autonomous vehicles by using a hierarchical RL framework.

The MARL works above typically make a control decision by choosing a next-hop node or edge (a.k.a intersection or road segment). Due to the well-known non-stationary issue in MARL, the asynchronicity in the MSD-SPP makes it harder to cooperate among vehicles if simply using the CTDE framework. Moreover, the next-hop node-based methods may lead to infinite loops due to incorrect decision-making. 

\section{Problem Definition and Overview}\label{sec3:def}
In this section, we first give our problem definition, then give an overview of our solution framework. Table \ref{tab:symbol} lists the definitions of the symbols.

 \subsection{Problem Definition}

\begin{definition}
\textbf{Road Network}. We define the road network as a directed graph $\mathcal G=(\mathcal V, \mathcal E)$, where $\mathcal V$ is a {set of nodes, with each node representing a road intersection} (we thus , and $ \mathcal E $ is an edge set. 

An edge $e_{ij} \in \mathcal E$ from $v_i$ to $v_j$ {indicates the road segment from the intersection $v_i$ to another one $v_j$}. The traffic capacity of road segment $e_{ij}$ is defined as $c_{ij}$. Denote the number of vehicles running on the road segment $e_{ij}$ as $n_{ij}$. If this number $n_{ij}$ exceeds $c_{ij}$, for simplicity, we assume that the travel speed along $e_{ij}$ is reduced to a smaller value by a fraction $\alpha \cdot \frac{c_{ij}}{n_{ij}}$, where $0<\alpha<1$. The greater the number $n_{ij}$ exceeds the capacity $c_{ij}$, the lower the travel speed along the road.
\end{definition}
 
 \begin{table}[htb] \footnotesize
 \caption{Summary of symbols}
     \begin{tabular}{l|l}
\hline
\textbf{Symbol} & \textbf{Definition} \\ \hline\hline
    $\mathcal G=(\mathcal V, \mathcal E)$ & \begin{tabular}[c]{@{}l@{}}a road network {with an intersection set $\mathcal V$ and road segment set $\mathcal E$}\end{tabular} \\
    $v_i$ , $e_{ij}$, $c_{ij}$ & intersection, road segment from  $v_i$ to $v_j$ and traffic capacity \\
    $L_l$ & the $l$-th vehicle {which travels from $v_l^s$ to $v_l^d$ at time $t_l^s$}\\
    $p_l$, $t_{L_l}$ & routing path and travel time of vehicle $L_l$ \\
    \hline
    $ M, L$ & No. of divided regions, and No. of vehicles \\
    $ R_i $, $ \pi^i$ & the $i$-th region and policy of $R_i$ \\ 
    $ \mathcal E_i^c $, $\mathcal V_i^c$ & cutting edges from $R_i$ and boundary nodes in $R_i$ \\
    \hline
    $ I_l $, $ H_l $ & connection graph and reachability graph of vehicle $L_l$ \\
    $q_t^{i,k}, \mathcal Q_t^i$ & the $k$-th request and request set received by $R_i$ at time $t$ \\
    $ o_t^{i,k}$, $ a_t^{i,k} $ & \begin{tabular}[c]{@{}l@{}}observ. and action of agent $R_i$ for $q_t^{i,k}$ at time $t$ \end{tabular} \\
    $o_t^{i,\mathcal G}$& road network observ. of agent $R_i$ at time $t$\\
    $o_t^{i,q_k}$ & routing request observ. of $q_t^{i,k}$\\
    $ s_t ^{i,k}$,$ r_t^{i,k} $ & \begin{tabular}[c]{@{}l@{}}global state and reward of agent $R_i$ for $q_t^{i,k}$ at time $t$ \end{tabular} \\
    $F, F', D_h$ & dim. of edge and request feature vectors, dim. of embeddings \\\hline
    $e_t^{i,\mathcal G}$& road network embedding of $R_i$ at time $t$ \\ 
    $e_t^{i, q_k},e_t^{i, \mathcal Q} $& embedding of request $q_t^{i,k}$ and all requests in $\mathcal Q_t^{i}$\\
    $u_t^{i,k} $ & scores on action roads of agent $R_i$ for $q_t^{i,k}$ at time $t$ \\ 
    $\mathcal D_i,\mathcal D$ & data buffer of agent $R_i$ and  centralized data buffer\\
    \hline
    $E,T$ & No. of episodes, and No. of simulation steps in a episode  \\
    $\theta_i,\phi$ & parameters of agent $R_i$'s policy and the critic \\
    $\hat G_t^{i,k},\hat A_t^{i,k}$ & discounted return and advantage of action $a_t^{i,k}$ \\
    \hline

\end{tabular}
\label{tab:symbol}     
\end{table}

\begin{definition}
\textbf{Vehicle}. Given $L$ vehicles travelling in a road network $\mathcal G$, each vehicle with a unique ID $l$ is represented with a triplet $L_l=\langle v^s_l, v^d_l, t^s_l \rangle $, where $v_l^s$ and $v_l^d$ represent the travel source and destination intersection nodes, respectively, while $t_l^s$ indicates the departure time. For simplicity, we assume that all vehicles are travelling at the same speed if no traffic congestion occurs. The routing path of the vehicle consists of a sequence of nodes $p_l=[v^s_l, ..., v^d_l]$. If $t^d_l$ is the arrival time at the destination $v^d_l$, the travel time $t_{L_l}=t^d_l-t^s_l$.
\end{definition}

\begin{problem}
\textbf{MSD-SPP}. 
Considering $L$ vehicles navigation through a road network $\mathcal G$, we formulate a route planning policy as an optimization problem. The objective is to minimize the average travel time of all vehicles while adhering to the traffic capacity constraint of $\mathcal G$.
\begin{equation}
    \min_{\mathbf{\pi}}\frac{1}{L}\sum_{l=1}^{L}t_{L_l}  
\end{equation}
Here, the traffic speed along the road segment $e_{ij}$ decreases and vehicles suffer from travel time penalties, if the number $n_{ij}$ of vehicles on $e_{ij}$ exceeds its capacity constraint $c_{ij}$.
\end{problem}

\subsection{Overall Framework}
\begin{figure*}[htp]
\centering\vspace{-2ex}   
\includegraphics[width=.7\linewidth]{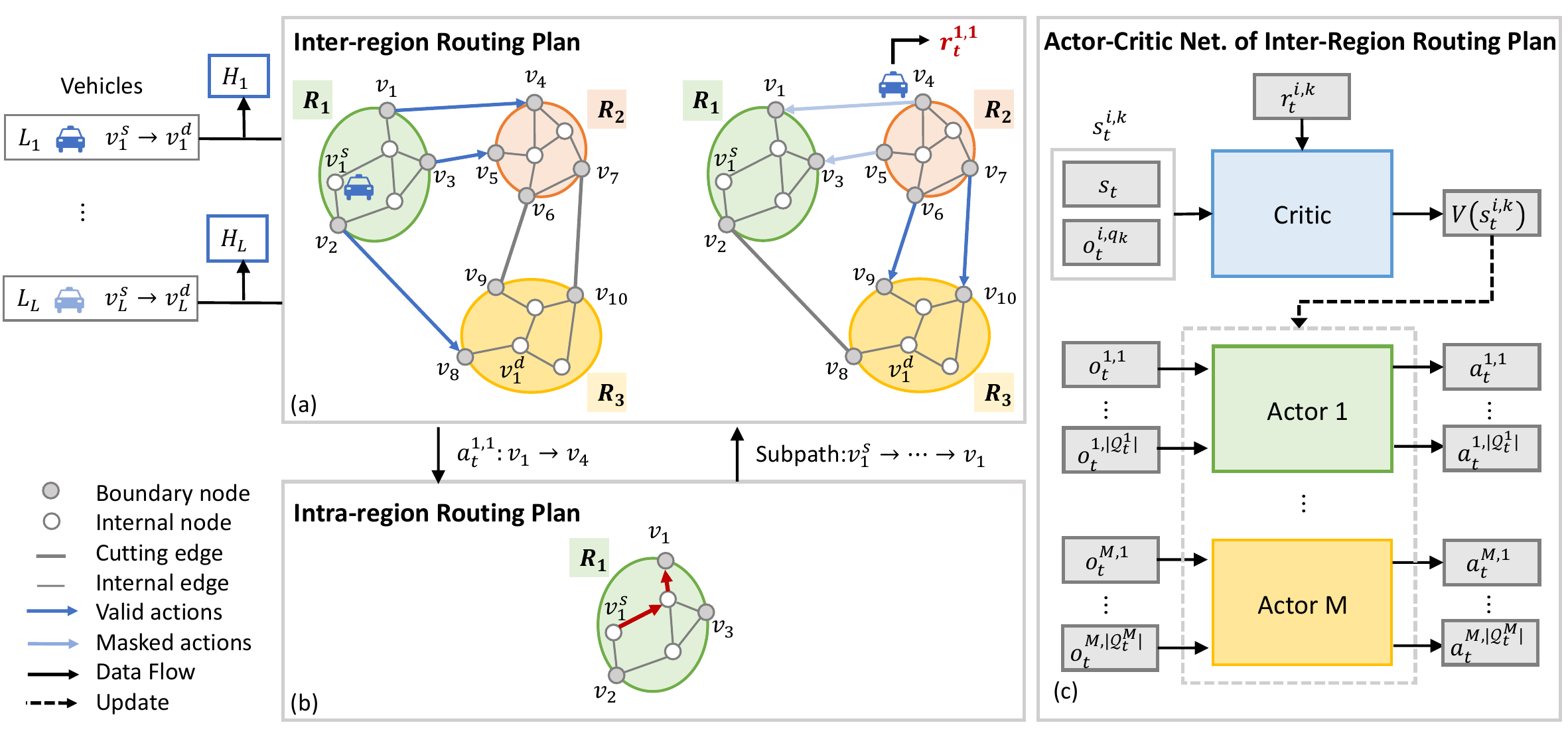}
\caption{Pipeline of the proposed method}
\label{fig:pipeline}
\centering
\end{figure*}

{Our proposed framework is illustrated in Figure \ref{fig:pipeline}. To begin with, we divide the road network into $M$ regions, and transform the vehicle routing from source to destination intersections into the routing from source to destination regions, followed by routing within the destination region towards the destination intersection. More specifically, we formulate MSD-SPP as a two stage process, including \emph{inter-region} and \emph{intra-region} planing. Since the number of regions is significantly smaller than the number of intersections, we expect that the computational cost of the two-stage process is much lower than the original one.}

In terms of road network division, we minimize the number of cutting edges across regions. For each region $R_i$, we denote $\mathcal E_i^c$ to be the \textit{set of cutting edges} originating from $R_i$, and $\mathcal V_i^c$ to be the \textit{set of boundary nodes} within the region $R_i$  that are connected by the cutting edges $\mathcal E_i^c$. In Figure \ref{fig:pipeline}(a), we have $M=3$ regions. The region $R_1$ involves three cutting edges $\mathcal E_1^c= \left \{ v_1 \to v_4, v_3 \to v_5, v_2\to v_8 \right \}$ and three boundary nodes $\mathcal V_1^c=\left\{ v_1, v_2, v_3\right \}$. We can exploit existing algorithms, such as the classic work METIS \cite{karypis1997metis}, to perform this graph division, and tune the number of divided regions mainly depending upon the capacity of region agents. By assuming that an agent can observe the entire region including at most $E_r$ edges, we can roughly compute the number of divided regions by $\lceil \frac{E}{E_r}\rceil$ where $E$ is the total number of edges in the input graph $\mathcal G$.

Given the $M$ regions, we then treat each region as an individual agent, and assume that each agent can observe three following information, (1) the \emph{static information of a road network}, such as the GPS coordinates of intersections and the lengths of road segments in the entire road network, (2) the \emph{detail information within the region} (such as the average speed and travel time of every road segment in the region), and (3) the \emph{estimation information of neighbouring regions} (i.e., the roughly estimated number of vehicles and average vehicle speed of such a neighbour).
Given the observation, the agent can compute the internal routes within the region and plan the routes across divided regions by the cooperation with other agents.
In Figure \ref{fig:pipeline}(a-b), when the vehicle $L_1$ starts at $v_1^s$ within the region $R_1$, the region agent first performs the inter-region plan to enter the neighbouring region $R_2$ via a selected cutting edge $v_1\to v_4$, and then computes the intra-region route from $v_1^s$ to $v_1$. When the vehicle enters the chosen region $R_2$, the region agent next chooses the cutting edge $v_7 \to v_{10}$ and plans the intra-region route within $R_2$. After the vehicle enters the region $R_3$, the associated agent finally plans the intra-region route within $R_3$ from $v_{10}$ to the destination $v_1^d$.

\emph{Inter-region route planning}: We formulate the inter-region route planning as a decentralized partially observable Markov decision process (Dec-POMDP).
\begin{itemize}
    \item \textbf{Agent}. We model each region as an individual agent and denote the set of $|\mathcal R|=M$ agents by $\mathcal R$. When a vehicle $l$ starts {a trip $L_l$ from a source node or arrives at a new region, it sends a plan request involving the trip $L_l$ to the agent of the corresponding region.}
    Here, each request is composed of the current node and destination node of the vehicle $l$. The agent then plans the inter-region route and sends the planning result back to the vehicle. 
    \item \textbf{Action}. Suppose that a region agent $R_i$ receives a set of plan requests $\mathcal Q_t^i$ sent by vehicles at time step $t$. For a certain request $q_t^{i,k}$, the agent $R_i$ makes an action $a_t^{i,k}$ to perform the inter-region routing plan to select next region to visit and the entry of the next region. 
    In practice, the agent $R_i$ selects one cutting edge from the set $\mathcal E_i^c$, and defines the action $a_t^{i,k}$ as an cutting edge.
    Given $M$ agents and {$\sum_{i=1}^{M} |\mathcal Q_t^i|$} requests, we have joint action $a_t=(a_t^{1,1}, \dots, a_t^{M, |\mathcal Q_t^M|})$. Here, the length of the joint action dynamically differs depending upon the time step $t$.

    \item \textbf{Observation}. Since we focus on the inter-region routing plan in this paper, the observation  $o_{t}^{i,k}$ of an agent $R_i$ involves the two following parts.
    Firstly, the \textit{road network observation} $o_t^{i, \mathcal G}$$\in$ $\mathbb R ^{|\mathcal E_i^c| \times F}$ consists of the number $|\mathcal E_i^c|$ of $F$-dimensional  feature vectors regarding the cutting edges $\mathcal E_i^c$. Each $F$-dimensional vector involves the features of a cutting edge (i.e., the coordinates of starting and ending nodes, length and travel time of this edge) and traffic estimation (i.e., the estimated number of vehicles and average vehicle speed) of the neighbour region to which the cutting edge is connected. 
    Secondly, the \textit{plan request observation} 
    $o_{t}^{i,\mathcal Q} \in \mathbb R^{|\mathcal Q_t^i| \times F'}$ indicates the observation of the requests $\mathcal Q_t^i$, where $F'$ is the dimensionality of an individual feature vector. That is, for each request $q_t^{i,k} \in \mathcal Q_t^i$, the associated observation $o_{t}^{i,q_k}\in \mathbb{R}^{F'}$ involves the following items, (1) the coordinates of the current intersection $v_c$ and final destination $v_d$ of the request $q_t^{i,k}$, (2) the least travel time from the source $v_c$ in the request $q_t^{i,k}$ to the $|\mathcal E_i^c|$ cutting edges $\mathcal E_i^c$ (computed by the detailed intra-region traffic information at time step $t$), and (3) the shortest path among those paths from the $|\mathcal E_i^c|$ cutting edges $\mathcal E_i^c$ to the destination $v_d$ in the request $q_t^{i,k}$ (on the static road network). 
    \item \textbf{Policy}. Given the observation $o_{t}^{i,k}$ above, the policy $\pi ^i$ gives a probability $\pi ^i(a_t^{i,k}|o_t^{i,k})$ that the agent $R_i$ makes an action $a_t^{i,k}\in \mathcal E_i^c$. We denote the joint policy of $M$ agents as $\mathbf{\pi}=(\pi^1, ..., \pi^M)$.
    \item \textbf{Reward}. We define the local reward of agent $R_i$ to make the action $a_t^{i,k}$ for the request $q_t^{i,k}$ by $r_{t}^{i,k}=-(t'-t)$. Here, $t'$ is the time step when the vehicle arrives at the end of the road segment chosen by the action $a_t^{i,k}$ or finally reaches the destination. In Figure \ref{fig:pipeline}(a), when a vehicle $L_1$ starts from the source $v_1^s$ and sends a plan request, the agent $R_1$ receives the request and then selects the cutting edge $v_1 \to v_{4}$ as the action. {The reward in this case} is the negative value of the travel time of this vehicle $L_1$ on the subpath $v_1^s \rightsquigarrow v_4$ intermediately through the region $R_1$. Next, when the region agent $R_2$ selects the cutting edge $v_7 \to v_{10}$ as an action, where $v_{10}$ is inside the destination region, we similarly compute the reward of the action $v_7 \to v_{10}$ as the negative travel time on the subpath $v_4\rightsquigarrow v_1^d$ passing through the edge $v_7 \to v_{10}$. Since higher travel time leads to a smaller reward,  the reward definition is consistent with the objective of the MSD-SPP problem to minimize the average travel time. 
\end{itemize}

\emph{Intra-region routing plan}: Since the intra-region routing plan does not involve multi-agent cooperation, we exploit our previous work \cite{yin2023learn} to find the shortest path within a region from the current node toward a chosen cutting edge. By assuming that the traffic details {such as traffic speed and travel time} of all road segments inside the region can be observed, the DRL model selects the next-hop intersection whenever a vehicle arrives at an intersection, based on current traffic.

\subsection{Overview of Inter-Region Routing Plan}
Given the Dec-POMDP formulation above, we exploit the Actor-Critic (AC) network \cite{actor-critic} to train the policy of the inter-region route planning model with the CTDE paradigm \cite{mappo}. In Figure \ref{fig:pipeline}(c), at each time step, every \emph{actor} $R_i$ with $1\leq i\leq M$ (i.e., an agent) interacts with the environment in both the training and execution phases, by taking the local observation $o_{t}^{i,k}$ as input and making an action $a_t^{i,k}$. Instead,  the centralized \emph{critic} (i.e., a neural network as a value estimation function) 
works only in the training phase. After receiving the reward $r_{t}^{i,k}$ for the action $a_t^{i,k}$, the critic evaluates the value of each individual state $s_t^{i,k}$ in order to determine whether the action $a_t^{i,k}$ made by agent $R_i$ is good or bad from a global view and next update the actor. Here, we develop a local state $s_t^{i,k}$ by including a global state $s_t$ and request observation $o_{t}^{i,q_k}$ (that will be  given very soon). Until now, in the execution phase, the agents (i.e., actors) perform in a fully decentralized manner to make actions and yet the centralized critic is unnecessary.

\textbf{Global State}: We define the global state $s_t$ at time step $t$ as the concatenation of the features regarding all cutting edges $\mathcal E^c$ in the road network $\mathcal G$. Recall that each actor (agent) $R_i$ makes an action to change the environment, leading to the traffic changes on cutting edges $\mathcal E_i^c$. Regarding the traffic on cutting edges, the global state $s_t$ is implicitly consistent with the joint observations and actions of $M$ region agents. Moreover, it is rather comfortable to learn the traffic on cutting edges as the global state $s_t \in \mathbb{R} ^{|\mathcal E^c|\cdot F}$, when compared to the explicit representation of joint actions and observations by $M$ agents in the asynchronous MSD-SPP setting. In this way, we exploit the low-dimensional global state to implicitly represent the high-dimensional joint observations and actions of multi-agents and ensure the stationarity of the MARL training in the asynchronous MSD-SPP setting.

\textbf{Local State}: When an agent $R_i$ takes an action $a_t^{i,k}$ {in response to} the request $q_t^{i,k}$, we define the local state $s_t^{i,k} = [s_t,o_t^{i,q_k} ] \in \mathbb R^{|\mathcal E^c|\cdot F+F'}$ as the concatenation of the global state $s_t$ above and the request observation $o_t^{i,q_k}$. Given this local state $s_t^{i,k}$, the critic estimate the state value $V(s_t^{i,k})$, which next updates the policy of the actor $R_i$. In this way, the critic evaluates the state value from a global view and ensures the cooperation in MARL.


To perform the cooperative routing plan in the asynchronous setting, we develop three components on top of the actor-critic (AC) MARL framework. (1) To address the non-stationary issue of MARL training in the asynchronous MSD-SPP setting, we design an \emph{asynchronous trajectory collection mechanism} (Section \ref{sec:critic}) in the critic network. That is, with the help of the developed global state above, we can asynchronously collect the trajectory of each agent, by inserting the transition of the agent into its separate data buffer immediately when it takes an action. Different from the synchronous trajectory collection manner, this method does not require that all agents take actions at each time step, and thus avoid the redundancy in the training trajectories. (2) To achieve vehicle cooperation, we develop a novel \emph{actor network} (Section \ref{sec:actor}). The actor effectively encodes road network traffics and routing requests to generate the probability distribution of choosing cutting edges (i.e., action space). In this way, for those vehicles with the same or close destinations, the actor then assigns a higher probability of choosing alternative cutting edges as actions. In this way, such vehicles have little chance to travel on the same road segments to avoid traffic congestion on these segments. (3) To tackle the infinite loop issue caused by incorrect routing decisions, we exploit a \textit{reachability graph} $H_l$ (Section \ref{sec:rg}) for each vehicle $L_l$ based on the static graph information. This graph eliminates road segments leading to infinite loops, and retains only those potential paths with short lengths. Then, we choose those cutting edges appearing within the reachability graph as valid actions.

\section{Solution Detail}\label{sec4:detail}
{In this section, we give the details of the three components (the critic, actor,  and reachability graph), and then describe the training method of our framework.}

\subsection{Critic Design} \label{sec:critic}

The key of our critic network is to develop asynchronous trajectory collection in the asynchronous MSD-SPP setting. We first give the general idea as follows. In the asynchronous MSD-SPP setting, some agents take actions and yet others not at the time step $t$. Whenever some actions are made, the change of the environment always occurs. Thus, we can exploit the developed global state to learn such a change and perform asynchronous trajectory collection as follows. 

More specifically, whenever an agent $R_i$ takes an action, it inserts its own trajectory data into a separate buffer $\mathcal D_i$. Denote a transition of a region agent $R_i$ as $(o_t^{i,k}, a_t^{i,k}, s_t^{i,k}, r_t^{i,k}, s_{t'}^{i', k'})$, i.e., given the local state $s_t^{i,k}$ and observation $o_t^{i,k}$ at time step $t$, the agent $R_i$ makes an action $a_t^{i,k}$ and changes the local state to $s_{t'}^{i', k'}$ at time step $t'$ with the reward $r_t^{i,k}$. 

\begin{figure}
\centering\vspace{-2ex}
\subfigure[Planning Actions and Trajectory Collection]{
\includegraphics[width=0.6\linewidth]{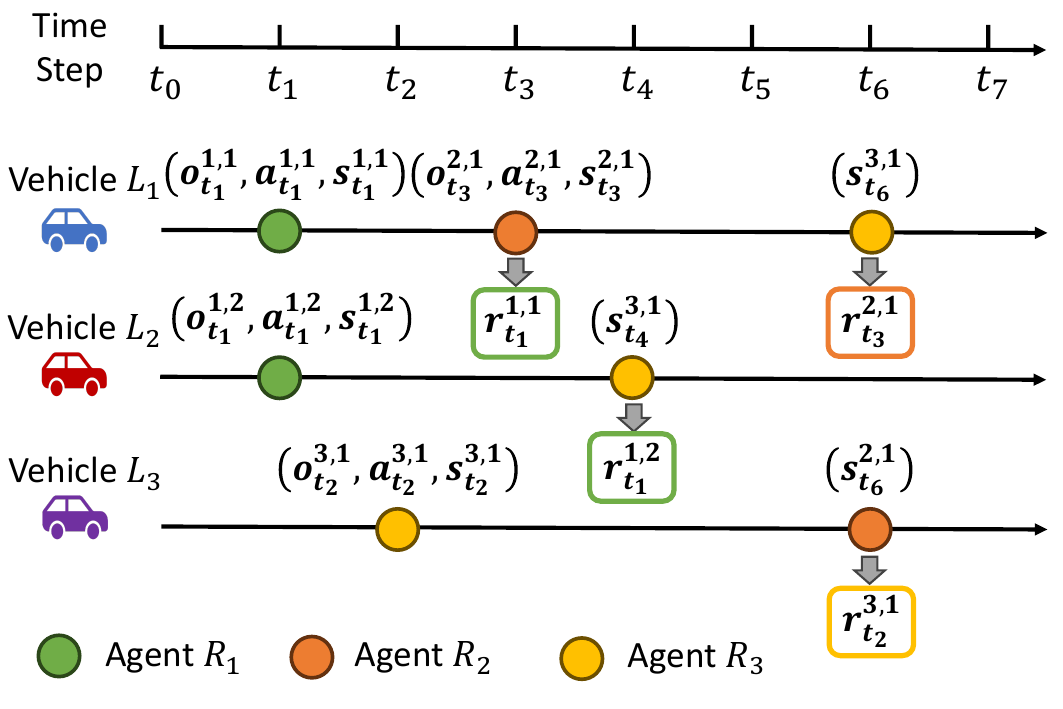}
}
\subfigure[Planning Paths]{
\includegraphics[width=0.35\linewidth]{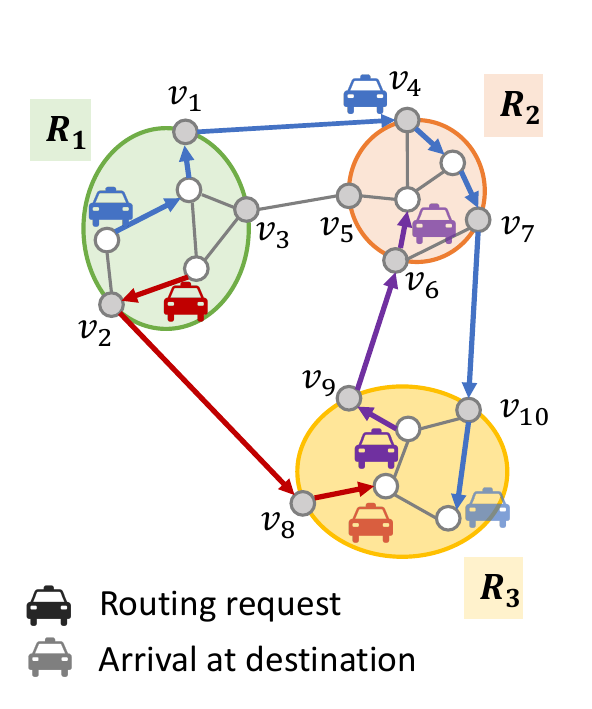}
}
\caption{Example for asynchronous trajectory collection}
\label{fig:asynchronous}
\centering
\end{figure}

Figure \ref{fig:asynchronous} gives an example of the asynchronous trajectory collection. At time step $t_1$, vehicles $L_1$ and $L_2$ start from region $R_1$. The region agent $R_1$ takes two routing actions for their requests, i.e, $a_{t_1}^{1,1}$ and $a_{t_1}^{1,2}$, respectively. Yet the agent $R_2$ and $R_3$ do not need to make an action at time step $t_1$. 
Here, the critic can construct the global state $s_{t_1}$ at time step $t_1$ and the local states $s_{t_1}^{1,1}=[s_{t_1}, o_{t_1}^{1,q_1}]$ and $s_{t_1}^{1,2}=[s_{t_1}, o_{t_1}^{1,q_2}]$ when the agent $R_1$ take actions $a_{t_1}^{1,1}$ and $a_{t_1}^{1,2}$ for the two vehicles $L_1$ and $L_2$, respectively. Regarding the reward of the action $a_{t_1}^{1,1}$, it is computed as $r_{t_1}^{1,1}= - (t_3 - t_1)$. Now we can insert the transition $(o_{t_1}^{1,1}, a_{t_1}^{1,1}, s_{t_1}^{1,1}, r_{t_1}^{1,1}, s_{t_3}^{2,1})$ into the buffer of agent $R_1$. After that, when the vehicle $L_1$ arrives at node $v_4$ at time step $t_3$, the agent $R_2$ selects the edge $v_7 \to v_{10}$ as the action based on the local observation $o_{t_3}^{2,1}$. Again, the critic can construct the global state $s_{t_3}$ and the local state $s_{t_3}^{2,1}=[s_{t_3}, o_{t_3}^{2,q_1}]$ when agent $R_2$ takes the action $a_{t_3}^{2,1}$. When the vehicle $L_1$ finally arrives at its destination in region $R_3$ at time step $t_6$, the critic can again compute the global state $s_{t_6}$, local state $s_{t_6}^{3,1}$, and the reward of its previous action $a_{t_3}^{2,1}$ as $r_{t_3}^{2,1}=-(t_6 - t_3)$. Similar to $L_1$, we can collect the transition trajectories of $L_2$ and $L_3$ as follows:
\begin{equation*}
\begin{array}{l}\small
 \mathcal D_1: [(o_{t_1}^{1,1}, a_{t_1}^{1,1}, s_{t_1}^{1,1}, r_{t_1}^{1,1}, s_{t_3}^{2,1}), (o_{t_1}^{1,2}, a_{t_1}^{1,2}, s_{t_1}^{1,2}, r_{t_2}^{1,2}, s_{t_4}^{3,1}) ] \\
 \mathcal D_2: [(o_{t_3}^{2,1}, a_{t_3}^{2,1}, s_{t_3}^{2,1}, r_{t_3}^{2,1}, s_{t_6}^{3,1}) ]\\
\mathcal D_3:[(o_{t_2}^{3,1}, a_{t_2}^{3,1}, s_{t_2}^{3,1}, r_{t_2}^{3,1}, s_{t_6}^{2,1}) ] 
\end{array}
\end{equation*}

After collecting the trajectory data of each agent, we can apply the CTDE paradigm to train the standard MARL model on the collected data. That is,
by using the global state $s_t$ and the request observation $o_{t}^{i,q_k}$ as the local state $s_t^{i, k}$ of agent $R_i$, the centralized critic exploits an MLP network to estimate the value of $s_t^{i, k}$, i.e., $V(s_t^{i,k})=\mathrm{MLP}([s_t, o_t^{i, q_k}])$. In this way, with the value $V(s_t^{i,k})$, the critic can evaluate how the action $a_t^{i,k}$ is good or bad to update the policy of the local actor $R_i$.

\subsection{Actor Design} \label{sec:actor}
\begin{figure}
\centering\vspace{-2ex}
\includegraphics[width=1.0\linewidth]{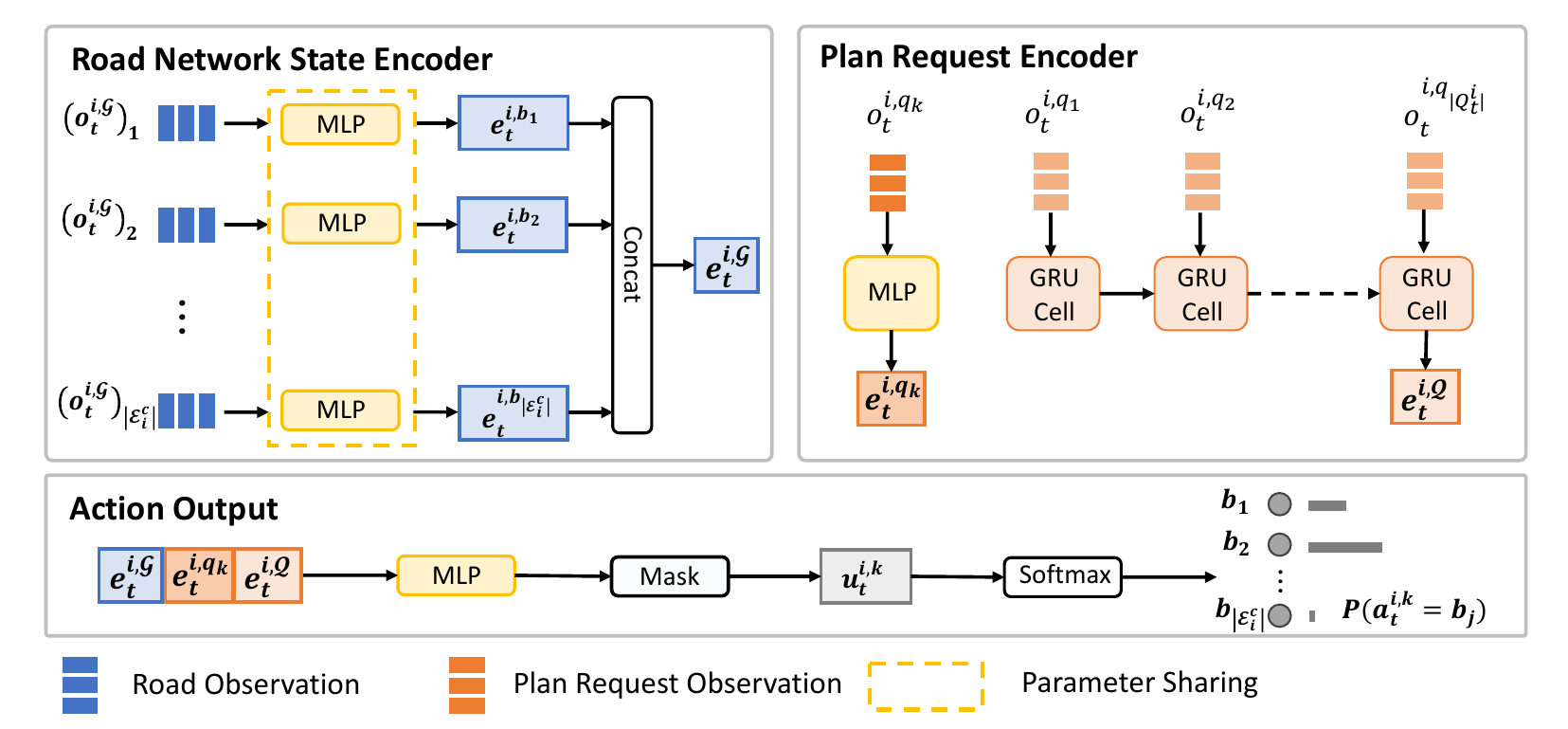}
\caption{Architecture of the actor}
\label{fig:actor}
\centering
\end{figure}


To facilitate vehicle cooperation and avoid congestion caused by planning the same path for vehicles sharing similar or proximate destinations, we propose a novel actor network in Figure \ref{fig:actor}, consisting of an encoder of the road network, an encoder of plan requests, and an output module. 

(1) The encoder of the road network learns the traffic features of the road network. Recall that the road network observation $o_t^{i,\mathcal G}$ involves the number $|E_i^c|$ of $F$-dimensional feature vectors regarding the cutting edges $\mathcal E_i^c $. Thus, for each cutting edge $b_j\in \mathcal E_i^c$, we denote its feature vector by $( o_t^{i, \mathcal{G}})_j$, and feed this vector into multi-layer perceptrons (MLP) to learn a latent embedding vector $ e_{t}^{i,b_j} $ for the edge $b_j$. By the concatenation of $|\mathcal E_i^c|$ vectors for the cutting edges $\mathcal E_i^c$, we have the road network state embedding $e_{t}^{i,\mathcal G}$.
\begin{equation} \small
e_{t}^{i,\mathcal G}=[e_{t}^{i,b_1},  \dots,  e_{t}^{i,b_{|\mathcal E_i^c|}} ],\mathrm{where} \; e_{t}^{i,b_j}=\mathrm{MLP} \left(( o_{t}^{i,\mathcal G})_j \right)
    \label{equ:road_encoder}
\end{equation}
where $[\cdot, \cdot]$ is the concatenate operation and $e_{t}^{i,b_j}\in \mathbb R^{D_h}, e_t^{i,\mathcal G} \in \mathbb R^{|\mathcal E_i^c| \times D_h}$, and $D_h$ is the dimensionality of the embeddings.

(2) The encoder of plan requests takes each request vector $o_{t}^{i,q_k}$ into an MLP and yields the embedding $e_{t}^{i,q_k} \in \mathbb R^{D_h}$ for the request $q_t^{i,k}$. To enable the cooperation among those plan requests $\mathcal Q_t^{i}$ within the region $R_i$, we learn an embedding for the entire plan requests. Since the number $|\mathcal Q_t^{i}|$ differs from the time step $t$, we employ a Gated Recurrent Unit (GRU) \cite{gru} cell to represent such requests $ \mathcal Q_t^i$ as follows. 
\begin{equation}
h_{k'}=\mathrm{GRU}(o_{t}^{i,q_{k'}},h_{k'-1}),k'=1,\dots,|\mathcal Q_{t}^{i}|
\label{equ:gru}
\end{equation}
where the request vector $o_t^{i,q_{k'}}\in \mathbb R ^{F'}$ and $h_{k'} \in \mathbb R ^ {D_h} $ are input vector and the hidden state of the GRU cell at step $k'$, respectively. Given the sequential GRU model, we take the hidden state of the final GRU cell as the embedding of $ \mathcal Q_t^i$, i.e., $e_{t}^{i,\mathcal Q}=h_{|\mathcal Q_t^i|} \in \mathbb R^{D_h}$.


(3) Now we have three embeddings, i.e., $e_{t}^{i,\mathcal G}$, $e_{t}^{i,q_k}$, and $ e_{t}^{i,\mathcal Q} $. With such embeddings as input, the actor produces a probability distribution on the cutting edges $\mathcal E_i^c$ and chooses the cutting edge with the highest probability. Specifically, the output module first maps the embeddings to a score vector $ u_t^{i,k} \in \mathbb R^{|\mathcal E_i^c|}$ on  cutting edges $\mathcal E_i^c$ with an MLP, {then employs a softmax function \cite{bishop2006pattern} to transform the edge scores into a probability distribution on $\mathcal E_i^c$. Here, a greater edge score corresponds to a higher probability. The intuition of the probability distribution is as follows. With help of the embeddings $e_t^{i,q_k}$ and $e_t^{i,\mathcal Q}$, the actor can identify those requests {towards the same or close} destinations. If the number of such requests is small, the actor tends to assign a high probability (near 1.0) to the cutting edge which leads to the shortest travel time for the request $q_t^{i,k}$. Otherwise, the actor assigns an even probability distribution across the cutting edges $\mathcal E_i^c$. In this way, we have chance to avoid planning too many vehicles on the same road segments and thus achieve cooperation among the vehicles .}

Specifically, we use a non-linear transformation implemented by MLP to achieve the mapping from embeddings to scores.
In addition, we apply the invalid action mask \cite{action_mask} with the help of the reachability graph (Section \ref{sec:rg}) to avoid loops in the solution paths. In this way, we compute the score of a cutting edge $b_j$ by $(u_t^{i,k})_j =  (  \mathrm{mlp} ( [ e_t^{i,\mathcal G}, e_t^{i, q_k}, e_t^{i, \mathcal Q} ] ) )_j$ if the chosen action $b_j$ is within the reachability graph (that will be given soon) and otherwise $(u_t^{i,k})_j = -\infty$. After that, we employ a softmax function \cite{bishop2006pattern} to compute the probability of selecting edge $b_j\in \mathcal E_i^c$ based on the scores,
\begin{equation} \small
P(a_t^{i,k}=b_j)=\frac{\exp ( ( u_t^{i,k} )_j )}{  {\textstyle \sum_{j'=1}^{|\mathcal E_i^c|} } \exp (( u_t^{i,k} )_{j'} ) } 
\end{equation}

With the softmax, a greater score regarding a cutting edge $b_j$ leads to a higher probability of being chosen as an action, and thus the probability to select invalid edge becomes zero.

\subsection{Reachability Graph} \label{sec:rg}
Due to the lack of a global view, the inter-region routing plan may suffer from the issue of infinite loops. To overcome this issue, we propose to construct a \emph{reachability graph} for each vehicle $L_l=\langle v_l^s, v_l^d , t_l^s\rangle$. Firstly, when a vehicle $L_l$ travels from the source $v_l^s$ to the destination $v_l^d$, the source region agent processes the plan request and constructs the reachability graph $H_l$ based on the static graph information. The construction of such a reachability graph by the source agent makes sense, because every region agent can observe the static information of an entire road network, such as the GPS coordinates of intersections and the lengths of road segments. After that, when the trip continues, the graph $H_l$ is then sent to a neighbor region agent, which can mask invalid actions that will lead to loops via the received graph $H_l$.

\begin{figure}
\centering\vspace{-2ex}
\includegraphics[width=.8\linewidth]{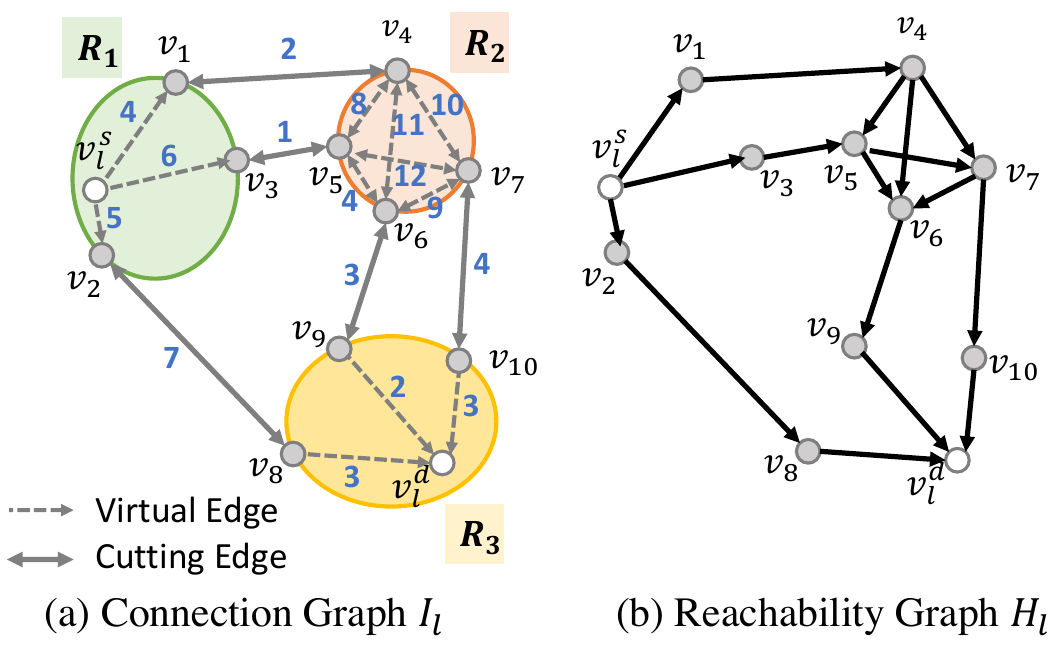}\vspace{-2ex}
\caption{An example of reachability graph construction }\vspace{-3ex}
\label{fig:dag}             
\centering
\end{figure}

To construct a reachability graph $H_l$, the source region first needs to build a \emph{connection graph} $I_l$ to represent the potential paths from the source $v_l^s$ to destination $v_l^d$ intermediately through boundary nodes and cutting edges. As shown in Figure \ref{fig:dag}(a), given the source $v_l^s$ and destination $v_l^d$, we construct three virtual edges from source node $v_l^s$ to the boundary nodes in source region $R_1$, $v_l^s \to v_1$, $ v_l^s \to v_2$ and $v_l^s \to v_3$. Again, we construct virtual edges from boundary nodes in destination region $R_3$ to the destination node $v_l^d$. Moreover, we construct virtual edges among boundary nodes in the intermediately connected regions. These virtual edges are weighted by the shortest path distances on the static road network.

Now, with the connection graph $I_l$, we can build a region-level path by {by a path on $I_l$}. For instance, the path $v_l^s \to v_1 \to v_4 \to v_7 \to v_{10} \to v_l^d$ indicates that the vehicle starts from region $R_1$, with the chosen cutting edge $v_1 \to v_4$ to enter the region $R_2$ and the final cutting edge $v_7 \to v_{10}$ to reach the destination region $R_3$.
Note that the connection graph $I_l$ may involve many loops, for example, $v_4 \to v_6 \to v_7 \to v_5 \to v_4$. To avoid loops during route planning, we convert the connection graph $I_l$ into the reachability graph $H_l$ by a directed acyclic graph (DAG) conversion algorithm \cite{gddr} with the purpose to {prune those edges that lead to loops} and retain relatively short paths as many as possible. Due to space limit, we refer interested readers to the paper \cite{gddr}. On the connection graph $I_l$ in Figure \ref{fig:dag}(a), we give the reachability graph $H_l$ in Figure \ref{fig:dag}(b) by using the conversion algorithm \cite{gddr}.

Since the size of the connection graph $I_l$ is much smaller than the original road network graph, the construction cost of $H_l$ is trivial. When a region agent handles the request from vehicle $L_l$, it will mask the invalid action roads that are infeasible in the graph $H_l$. For instance, the reachability graph $H_l$ indicates that agent $R_2$ receives the request from the vehicle $L_l$, and the valid actions for its request are confined to $\{v_6 \to v_9, v_7 \to v_{10} \}$.

\subsection{Training Step} \label{sec:train}
Given the trajectory collection in Section \ref{sec:critic}, we train the developed actor-critic network with the Multi-agent Proximal Policy Gradient (MAPPO) \cite{mappo} on the data buffers of $M$ agents. Algorithm \ref{alg:train} gives the training steps of asyn-MARL.

\begin{algorithm}[htbp]	 \small
\SetKwInOut{Input}{input}\SetKwInOut{Output}{output}
Initialize the params. of $M$ actors $\theta_i$ and centralized critic $\phi$;\\
\For {episode =1 to $E$}{
Set data buffer $\mathcal D_i=\{\}$ for $i=1 \dots M$; \\
    \For {time step $t\in{1 \dots T}$}{
        Get the global network state $s_t$; \\
        \For {$1\leq i \leq M$}{
            Get the plan request set $\mathcal Q_{t}^i$ of actor $\theta_i$; \\
            \For {request $q_t^{i,k} \in \mathcal Q_{t}^i$}{
            $R_{i}$ makes action $a_{t}^{i,k}$ on obs. $o_{t}^{i,k}$ with $\pi^i$; \\
            Insert $(o_{t}^{i,k}, a_{t}^{i,k}, s_{t}^{i,k}, r_{t}^{i,k}, s_{t'}^{i', k'}) $ into $\mathcal D_i$;\\
            Compute advantage $\hat A_t^{i,k}$ of $a_t^{i,k}$ by Eq.\ref{equ:adv};\\
            Compute discounted return $\hat G_t^{i,k}$ by Eq.\ref{equ:return};
            }
        }
    }
    \For {$i=1, \dots, M$}{
    Update the actor parameter $\theta_i$ by Eq. \ref{equ:ppo} on $\mathcal D_i$; \\}
    $\mathcal D = \textstyle \bigcup_{i=1}^{M}\mathcal D_i $; \\
    Update the critic parameter $\phi$ by Eq. \ref{equ:value} on $\mathcal D$; \\
    
}
\caption{Training the AC network of asyn-MARL}
\label{alg:train}
\end{algorithm} 

In line 1, the algorithm initializes the network parameters of $M$ actors and the critic, i.e., $\theta_i$ with $1\leq i\leq M$ and $\phi$ randomly. Next, in each episode, we first empty the data buffers of all actors (line 3). 
At each time step, the critic accesses the global state $s_t$ (line 5). Each actor receives a set of routing requests (line 7), and makes actions with its policy $\pi^i$, which is parameterized with $\theta_i$, for the observation $o_{t}^{i,k}$ (line 9). Then, the actor receives the reward $o_{t}^{i,k}$ and inserts the transition into its data buffer (line 10). Each transition consists of the local observation $o_t^{i,k}$, action $a_t^{i,k}$, reward $r_t^{i,k}$, state $s_t^{i,k}$, and the new state $s_{t'}^{i', k'}$ after the action is made. After that, we compute the advantage $\hat{A}_{t}^{i,k}$ by Equation \ref{equ:adv} and the discounted return $\hat{G_t^{i,k}}$ by Equation \ref{equ:return}. Here, $\hat{A}_{t}^{i,k}$ and $\hat{G_t^{i,k}}$ are used to update the parameters of actors and critic.
\begin{equation} 
\hat{A}_{t}^{i,k}=r_t^{i,k} + \gamma V(s_{t'}^{i',k'}) -V(s_t^{i,k}) 
\label{equ:adv}
\end{equation}
where $\gamma \in (0,1]$ is the discount factor and $V$ is the global value function of the centralized critic (here, the value function of the critic is parameterized with $\phi$).
\begin{equation}
\hat{G_t^{i,k}} = r_t^{i,k}+\gamma V(s_{t'}^{i',k'})
    \label{equ:return}
\end{equation}
where $r_t^{i,k}$ is the reward for action $a_t^{i,k}$ and $V(s_{t'}^{i',k'})$ is the value of the new state $s_{t'}^{i',k'}$ after the action $a_t^{i,k}$ is made.

At the end of each episode, we update the policy parameters of each actor $\theta_i$ (line 17). By the PPO-Clip algorithm \cite{ppo}, we update the actor policy by maximizing the surrogate objective.
\begin{equation} \small
L(\theta_i) = \mathbb{E}_{o_t^{i,k}, a_t^{i,k}} [ \min  ( \rho_t^{i,k} \hat{A}_t^{i,k}, \mathrm{clip} ( \rho_t^{i,k}, 1-\epsilon, 1+\epsilon)  \hat{A}_t^{i,k} )  ] 
\label{equ:ppo}
\end{equation}
where $  \rho_t^{i,k} =\frac{\pi ^i(a_t^{i,k}|o_t^{i,k})}{\pi_{old}^i(a_t^{i,k}|o_t^{i,k})}$ is the ratio between the current policy $\pi^i$ and the old one $\pi_{old}^i$ that is used to sample the trajectory by the actor $\theta_i$. This ratio is clipped to $[1-\epsilon, 1+\epsilon]$ to avoid a significant policy update and $\epsilon \in (0,1)$ is the PPO clipping parameter.

Finally, in lines 19-20, we combine the $M$ data buffers to an entire buffer $\mathcal D$ and update the critic parameters $\phi$ by minimizing the loss $L(\phi)$, e.g., a mean-squared error, on the data buffer $\mathcal D$.
\begin{equation} 
L(\phi)=\mathbb{E}_{s_t^{i,k}}\left ( V(s_t^{i,k})-\hat{G}_t^{i,k} \right )^2 
    \label{equ:value}
\end{equation}

\section{Experiment}\label{sec5:experiment}
\subsection{Data Sets}
We conduct experiments on two data sets, a synthetic road network and a real-world road network, and perform traffic simulation on a widely used simulator, namely the Simulation of Urban Mobility (SUMO) \cite{sumo}.

\begin{figure}
\centering
\subfigure[Synthetic (100 nodes)]{
    \includegraphics[width=4.0cm]{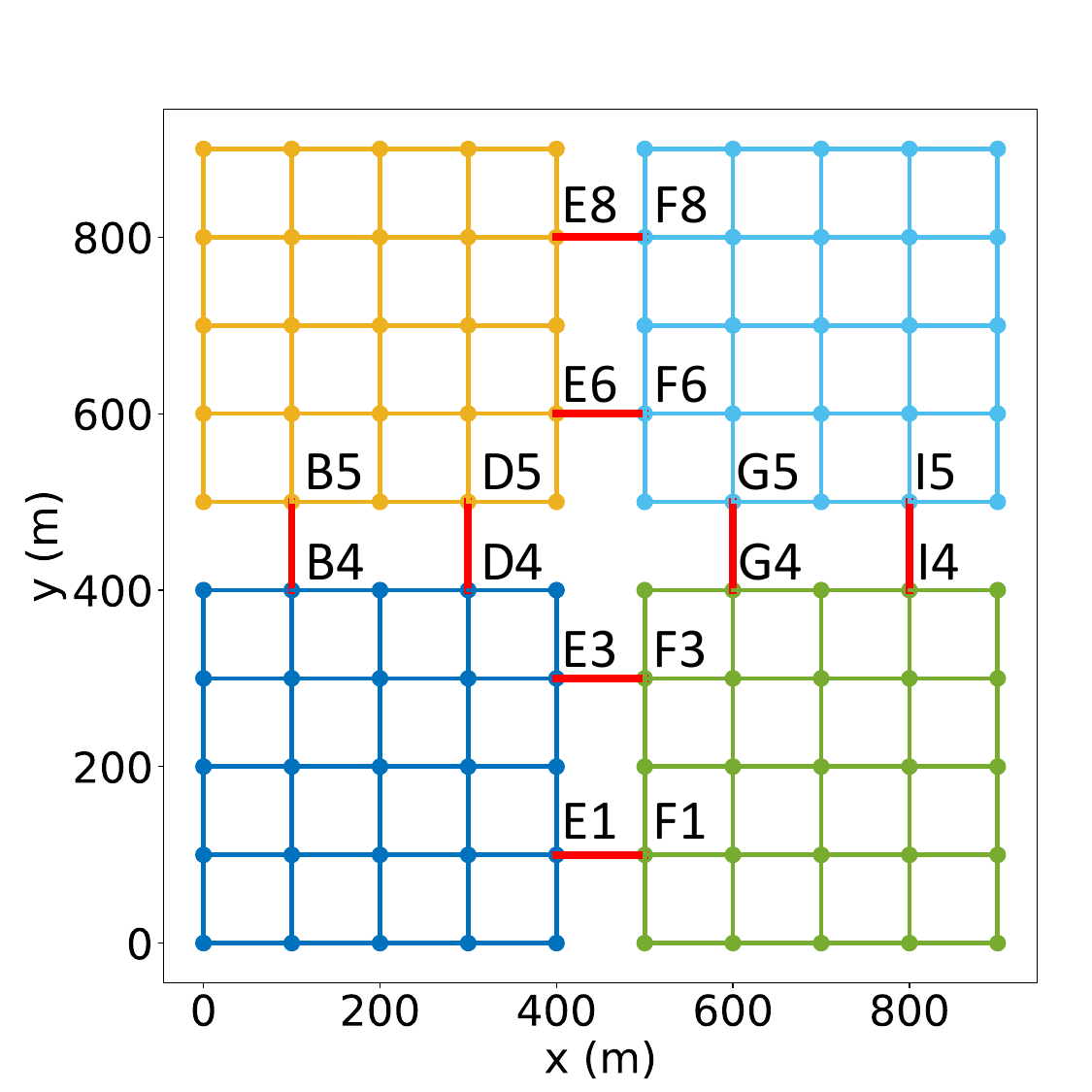}
}
\subfigure[Koln (2515 nodes)]{
    \includegraphics[width=4.0cm]{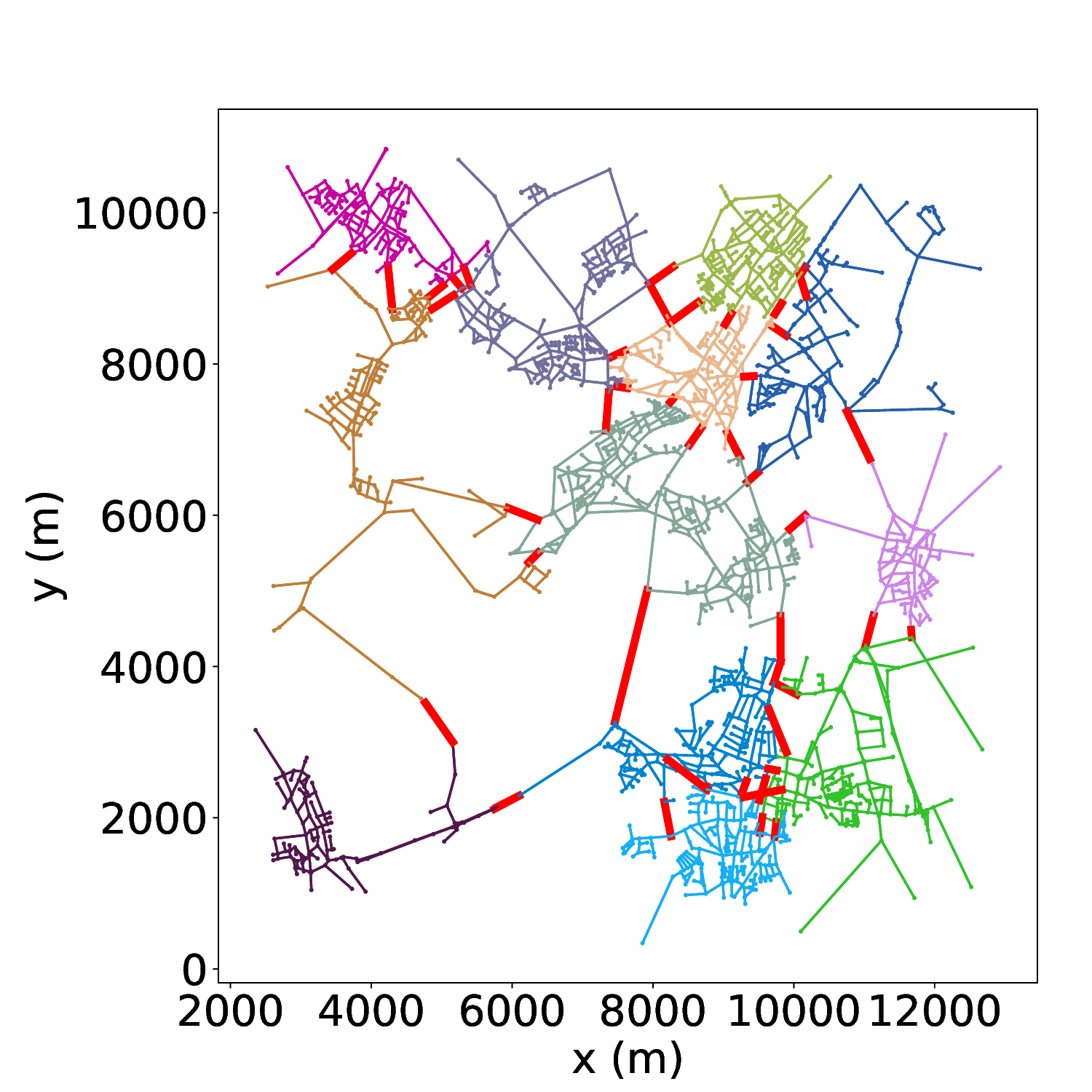}
}\vspace{-2ex}
\caption{Road network data sets (Cutting edges are plotted as red lines.)}
\label{fig:road_network}
\end{figure}

\textbf{(1) Synthetic data set}: Figure \ref{fig:road_network}(a) illustrates the synthetic road network of 4 regions, and each region consists of 25 intersections and 84 road segments. The length of road segments is 100 meters. Each region agent involves 4 cutting edges (i.e., the action space is 4), and the four regions totally 16 cutting edges. We generate traffic flows from the yellow region to the green one, and then randomly choose source and destination intersection nodes within the two regions, respectively. Moreover, we set the maximum speed on each road segment as 13.89 m/s.

\textbf{(2) Real data set}: We use a traffic data set provided by the previous work \cite{tmc14}, which provides the road network data in Koln, Germany. In this data set, we select an area of 100 $km^2$, from $6.166^{\circ}E$, $50.735^{\circ}N$ to $7.295^{\circ}E$,  $51.840^{\circ}N$. The selected area contains 2515 intersections and 5784 road segments. Following the data pre-processing technique  \cite{ahmed2016partitioning}, we remove the isolated road segments and dead-ends and then employ METIS \cite{karypis1997metis} to divide the road network into 12 regions with similar sizes and totally 84 cutting edges. As shown in Figure \ref{fig:road_network}(b), the number of cutting edges per a region ranges from 2 to 11 with an average 7. The road network involves 13 types of road segments. Depending upon the road types, we set the maximum travel speed of each road segment. For example, the maximum speeds of the primary road and secondary road are 19.44 m/s and 13.89 m/s, respectively.

\subsection{Traffic Simulation}
During the simulation, we inject vehicles into a road network every second until the total number of injected vehicles reaches a predefined value (e.g., 200 and 1500 for the synthetic and real road networks, respectively). When the number of vehicles on a road segment exceeds the road capacity limit, we follow the previous work \cite{sigspatial22} to simulate traffic congestion by decreasing the maximum speed by $\alpha * V_{max}$. Here, the factor $\alpha=0.1$ and $V_{max}$ is the original maximum speed of the road segment.

By default, we set the capacity limits on the synthetic network as 10, and Section \ref{sec:sensitivity} will evaluate the performance of asyn-MARL under various capacity limits and maximum traffic volumes on the synthetic network. Instead,  mainly due to the rather complex road types, we set a relatively high capacity limit 50 to the koln road network. The episode length of experiments is set as 600 seconds and 300 seconds for the two data sets. Table \ref{tab:sim_parameters} and \ref{tab:parameters} gives the simulation parameters on the two data sets and the hyper-parameters of the RL model, respectively.

\begin{table}\footnotesize 
\caption{Simulation Parameters for the two data sets}\label{tab2}
\vspace{-2ex}
\centering
\begin{tabular}{l|ll} \hline
\textbf{Data Set}                             & \textbf{Synthetic} & \textbf{Koln}  \\ \hline \hline
num. of Episode Steps $T$            & 600       & 300   \\
num. of Maximum Vehicles in Networks & 200       & 1500  \\
Road Capacity Limit                  & 10        & 50    \\
Average Road Maximum Speed (m/s)             & 13.89     & 11.42  \\ \hline
\end{tabular}
\label{tab:sim_parameters}
\end{table}

\begin{table} \footnotesize 
\caption{Parameter configuration for asyn-MARL}\vspace{-2ex}
\centering
\begin{tabular}{ll|ll}\hline
\textbf{Parameter} & \textbf{Value} & \textbf{Parameter}& \textbf{Value} \\ \hline \hline
dim. of embeddings $D_h$   & 32 & num. of mini-batches & 1    \\
PPO clipping parameter $\epsilon$ & 0.2  &num. of PPO epochs & 15   \\
discount factor $\gamma$ & 0.99   &actor learning rate & $10^{-5}$\\
maximum episode $E$ & 200  &critic learning rate &  $10^{-5}$   \\ \hline
\end{tabular}
\label{tab:parameters}
\end{table}

\subsection{Baselines and Metrics}
We compare our asyn-MARL with three traditional shortest path-based and two MARL-based algorithms.
\begin{itemize}
    \item \textbf{Random}: The agents perform the inter-region routing plan by randomly choosing a cutting edge to neighbour regions.
    \item \textbf{Shortest Path (SP)}: The agents select a cutting edge which appears on the shortest path to the destination based on static road networks.
    \item \textbf{Shortest Path First with Re-routing (SPFR)}: Following the previous work \cite{sigspatial22}, we pre-compute shortest paths on static road networks, and then update the route based on current traffic observation when a vehicle enters a new road segment. For fairness, we follow the same assumption as ours to ensure that this algorithm works in our setting: an agent can observe the detailed traffics on the road segments within the associated region, and roughly estimate the traffics of neighbour regions. 
    \item \textbf{Adaptive navigation (AN)}: The work \cite{sigspatial22} gives a MARL model \cite{sigspatial22} to train intersection agents by the deep Q-Network. Different from Q-routing, this work exploits the graph attention network \cite{gat} to learn the traffic on neighbour road segments of the intersection. 
    \item \textbf{Q-routing}: To be consistent with \cite{sigspatial22}, we choose Q-Routing \cite{qrouting} as our baseline. Though originally developed for network packet routing,  Q-Routing \cite{qrouting}  and our work share the very similar task to find the shortest path route. It estimates a Q-function $Q_x(d,y)$ with a large table to represent the time to arrive at node $d$ from a node $x$ by the way of $x$'s neighbor node $y$. We use the MARL implementation of Q-Routing provided by \cite{sigspatial22} to estimate the Q-function. This implementation treats each intersection as an agent and learns the values of selecting its neighbor nodes for a given destination node. This baseline is not aware of the global road network state.    
\end{itemize}
We use three metrics for performance evaluation.
\begin{itemize}
\item{\textbf{Throughput}}: the total number of vehicles which can successfully reach their destinations within an episode.
\item{\textbf{\underline{Av}erage \underline{t}ravel \underline{t}ime (AVTT)}}: For those vehicles which successfully reach their destinations within an episode, we compute the average travel time of such vehicles from the sources to destinations. 
\item{\textbf{Average CO2 emission volume (CO2)}}: Following the emission model HBEFA\footnote{\href{https://www.hbefa.net}{www.hbefa.net}}, we compute the average CO2 emission volume of each vehicle on the road network.
\end{itemize}

\begin{table*}[ht] \footnotesize  
\centering
\caption{Performance comparison on throughput, the average travel time (AVTT) and the average CO2 emission volume (ACE). (Results of Q-Routing and AN on Koln Data Set are omitted because they cannot scale to the large graphs)}\vspace{-2ex}
\label{tab:baseline}
\begin{tabular}{l|ccc|ccc}
\hline
  & \multicolumn{3}{c|}{\textbf{Sythetic Data Set}} & \multicolumn{3}{c}{\textbf{Koln Data Set}} \\ 
    & \textbf{Throughput}    & \textbf{AVTT} (s)    & \textbf{CO2} (kg)   & \textbf{Throughput}    & \textbf{AVTT} (s)    & \textbf{CO2} (kg)  \\ \hline\hline
Random    & 320.30        & 313.503    & 0.807     & 77.84       & 1580.444  & 3.352   \\ 
SP        & 738.02        & 158.837    & 0.416     & 268.08      & 1155.230  & 2.417   \\ 
SPFR      & 784.34        & 150.864    & 0.384     & 272.70      & 1113.291  & 2.385   \\ 
Q-Routing & 222.48        & 413.444    & 1.089     & /           & /         & /       \\ 
AN        & 764.15        & 168.967    & 0.444     & /           & /         & /       \\ \hline
asyn-MARL  & \underline{936.94}        & \underline{125.551}    & \underline{0.352}     & \underline{313.52}      & \underline{1090.434}  & \underline{2.489}   \\ \hline
\end{tabular}
\end{table*}

\subsection{Experiment Results}
\subsubsection{Baseline Study} \label{sec:baseline}

Table \ref{tab:baseline} gives the baseline result of our work (asyn-MARL) and five counterparts. From this table, we have the following findings.   On the two road networks, our work outperforms all five counterparts on the three metrics. Firstly, the two algorithms, random and SP, compute the static shortest paths and cannot adapt to the dynamic traffics caused by the competition of multiple routing plans. Thus, some road segments are with too many vehicles, leading to traffic congestion and slow travel time. For example, the average travel time of SP is 26$\%$ longer than asyn-MARL.

Secondly, SPFR updates the pre-computed shortest paths based on current road conditions and outperforms the Random and SP algorithms. However, it cannot offer vehicle cooperation and still lead to worse result than ours. 

Thirdly, the Q-Routing and AN algorithms exploit MARL to select next-hop intersections as actions. Here, since Q-routing is not aware of the global network state, it frequently makes incorrect decisions, leading to infinite loops. The infinite loops issue becomes greatly severe on large network graphs. Thus, some vehicles cannot reach their destinations, resulting in significantly longer travel time. In addition, AN leads to better performance than Q-Routing, due to the adopted graph attention network to incorporates the information of neighbour road segments. 

When comparing the results on two road networks, we find that the performance in the Koln data is worse than the one in the synthetic data, e.g., lower throughput and higher average travel time (AVTT) and CO2 emission. {The main reason is that} the area of the Koln network is almost 10 times greater than the synthetic data set, resulting in considerably longer travel time and CO2 emission volumes for vehicles to reach destinations. In terms of throughput, as shown in Table \ref{tab2},  the Koln data set is with a shorter  episode length and rather complex road segments than the synthetic data set, the associated 
  average travel speed is much slower, and more travel requests are not finished within an episode of the simulation, indicating smaller throughput on the Koln network. Moreover, the Koln data is with 2515 intersections. It is rather hard to train a very large MARL-based planning model with 2515 agents in Q-Routing and AN. Such result alternatively demonstrates the efficiency of our work asyn-MARL.
    



\subsubsection{Ablation Study} \label{sec:ablation}
To study the benefits of the three components, i.e., the actor network, the centralized critic and the reachability graph, we conduct ablation study and evaluate the performance of asyn-MARL with three following variants.
\begin{itemize}
    \item{\textbf{asyn-MARL w/o request embedding}: we remove the request encoding of other requests and output the action selection with simple MLP network directly.}
    \item{\textbf{IPPO}: we train the multi-agents independently with the PPO algorithm \cite{ppo}. Unlike our work asyn-MARL, this variant does not have a centralized critic, and instead each agent has a local version of the actor and critic by using the local observation as input yet without considering the joint observations and actions of other agents (i.e., the global state $s_t$ in asyn-MARL).} 
    \item{\textbf{asyn-MARL w/o reachability graph}: we remove the reachability graph and associated action masks during action decision.}
\end{itemize}
\begin{figure}[htb]
\centering
    \subfigure[Throughput]{
    \includegraphics[width=0.47\linewidth]{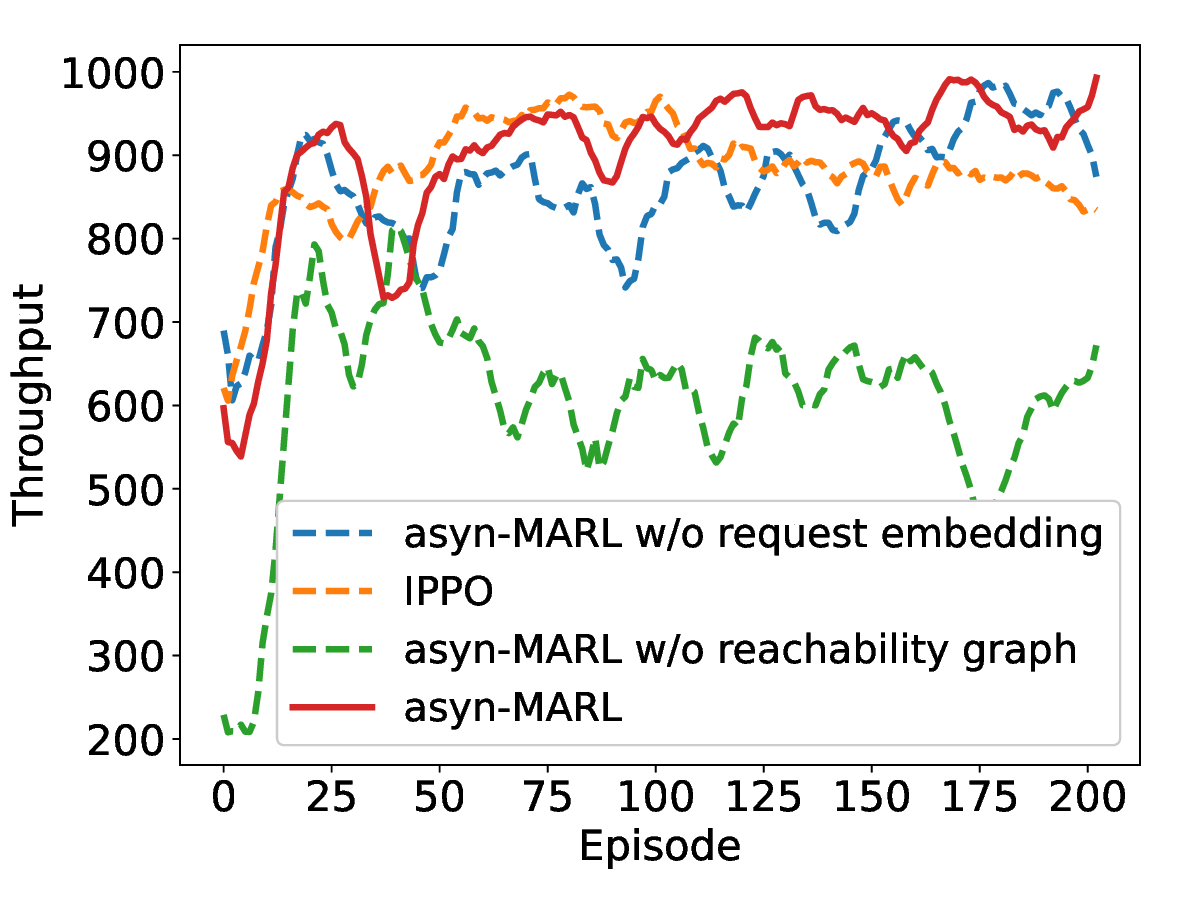}}
    \subfigure[Average Travel Time (second)]{
    \includegraphics[width=0.47\linewidth]{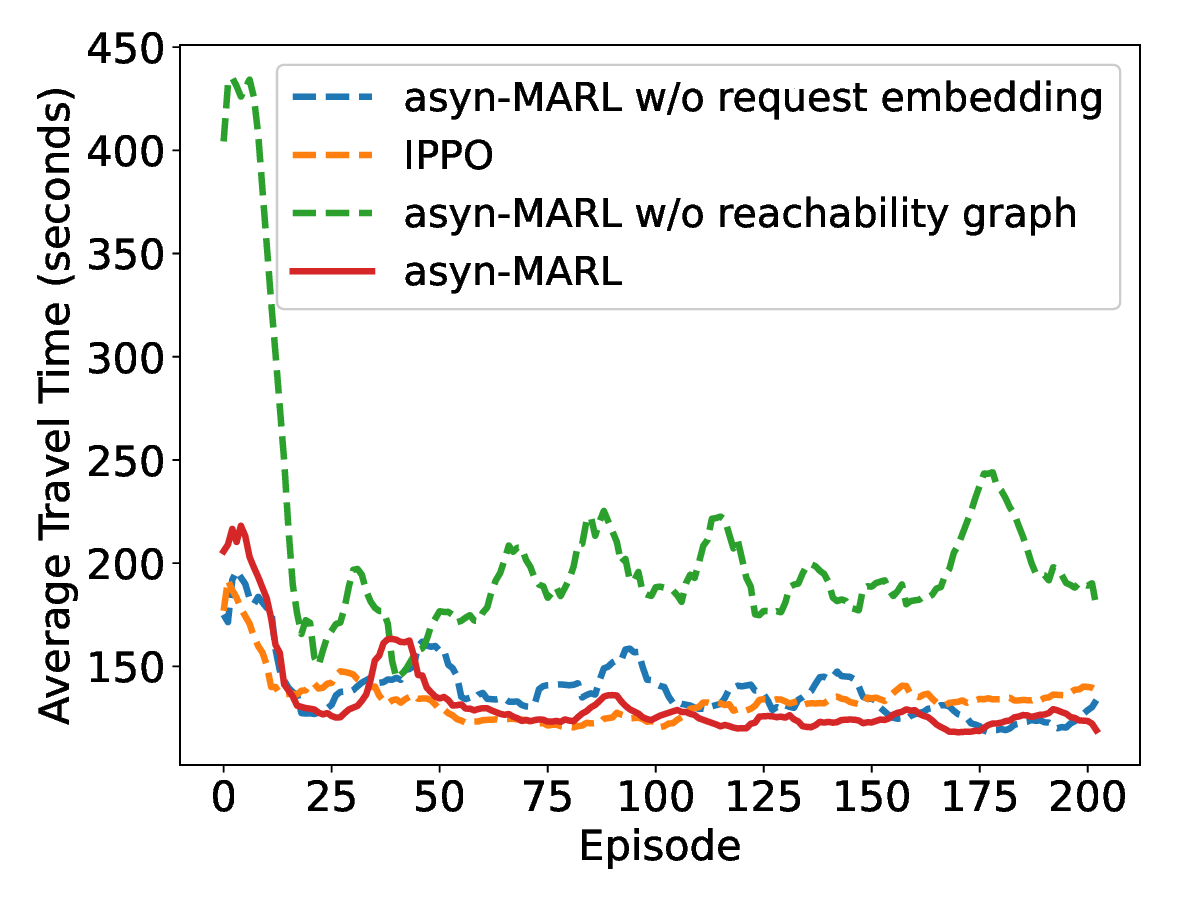}}\vspace{-2ex}
\caption{Results of Ablation Study}
\label{fig:ablation}
\end{figure}

As shown in Figure \ref{fig:ablation}, we plot the throughput and average travel time of the four algorithms during the training phase. Compared to our work asyn-MARL, the three variants lead to lower throughput and greater AVTT. In particular, the variant with no reachability graphs suffers from the worst performance. From this ablation study, the evaluation results indicate the importance of the three components in our framework aysn-MARL.

\subsubsection{Load Balance Study} \label{sec:load_balance}

\begin{figure*}[htbp]
    \centering
    \subfigure[Random]{
    \includegraphics[width=0.15\linewidth]{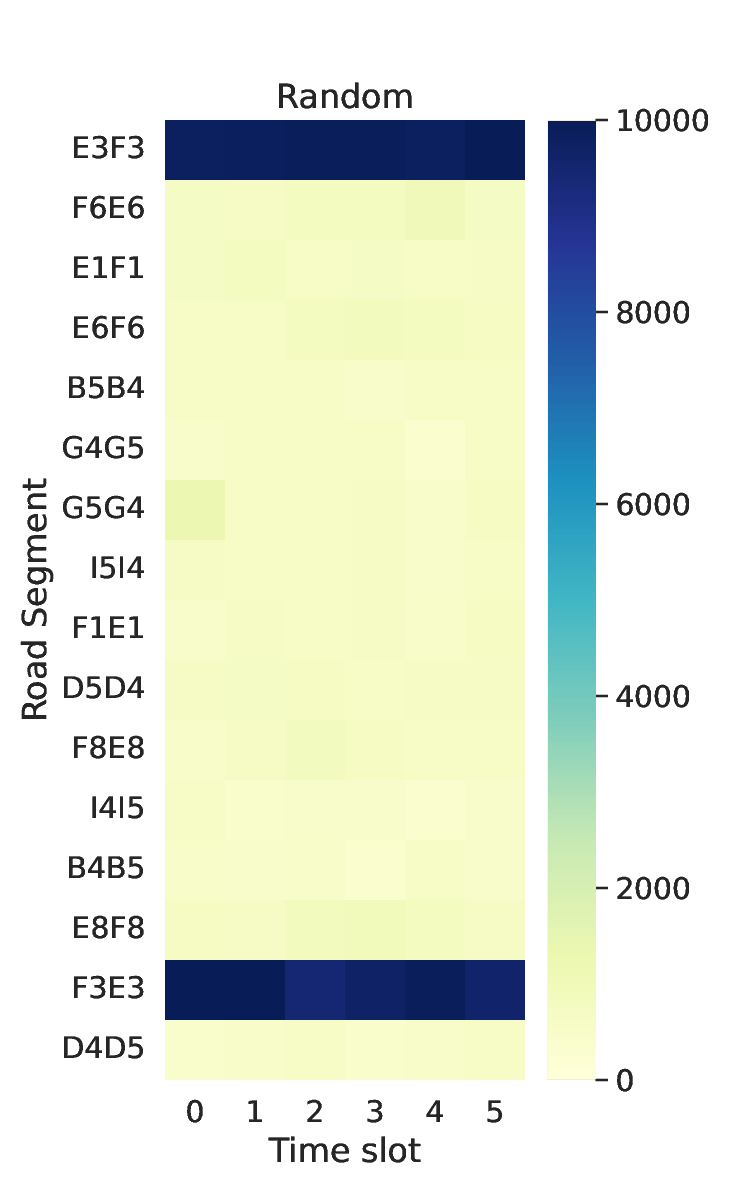}}
    \subfigure[SP]{
    \includegraphics[width=0.15\linewidth]{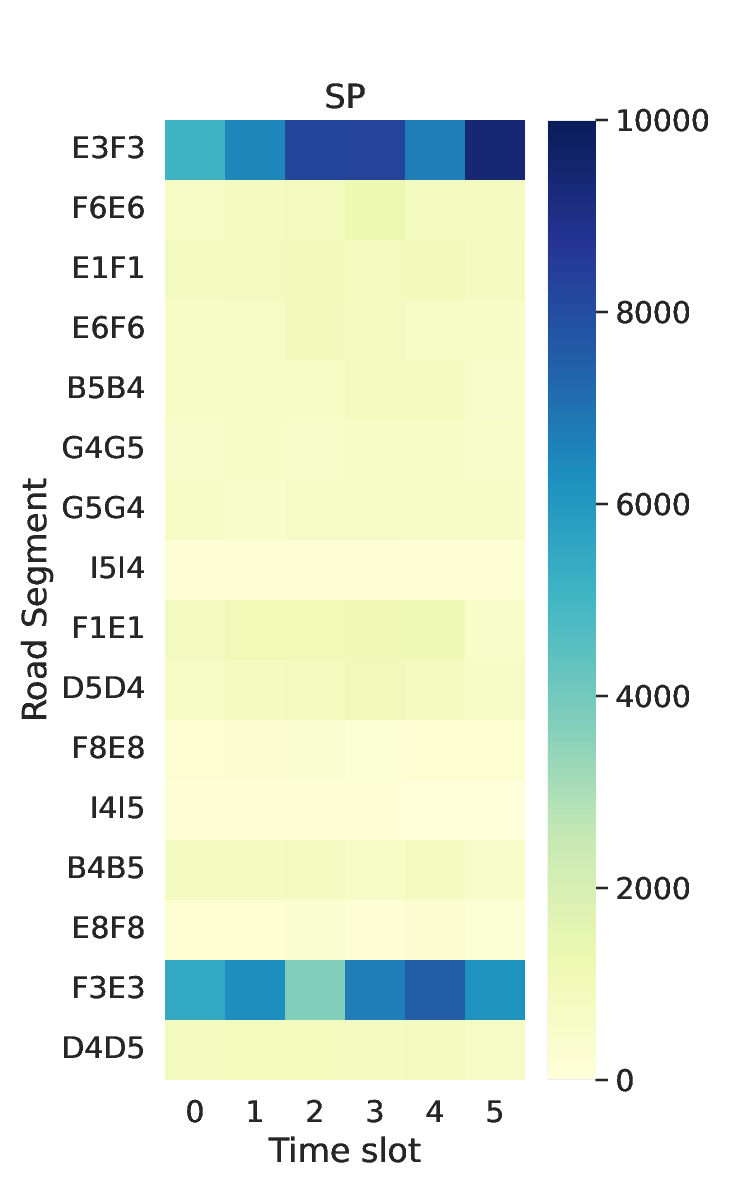}}
    \subfigure[SPFR]{
    \includegraphics[width=0.15\linewidth]{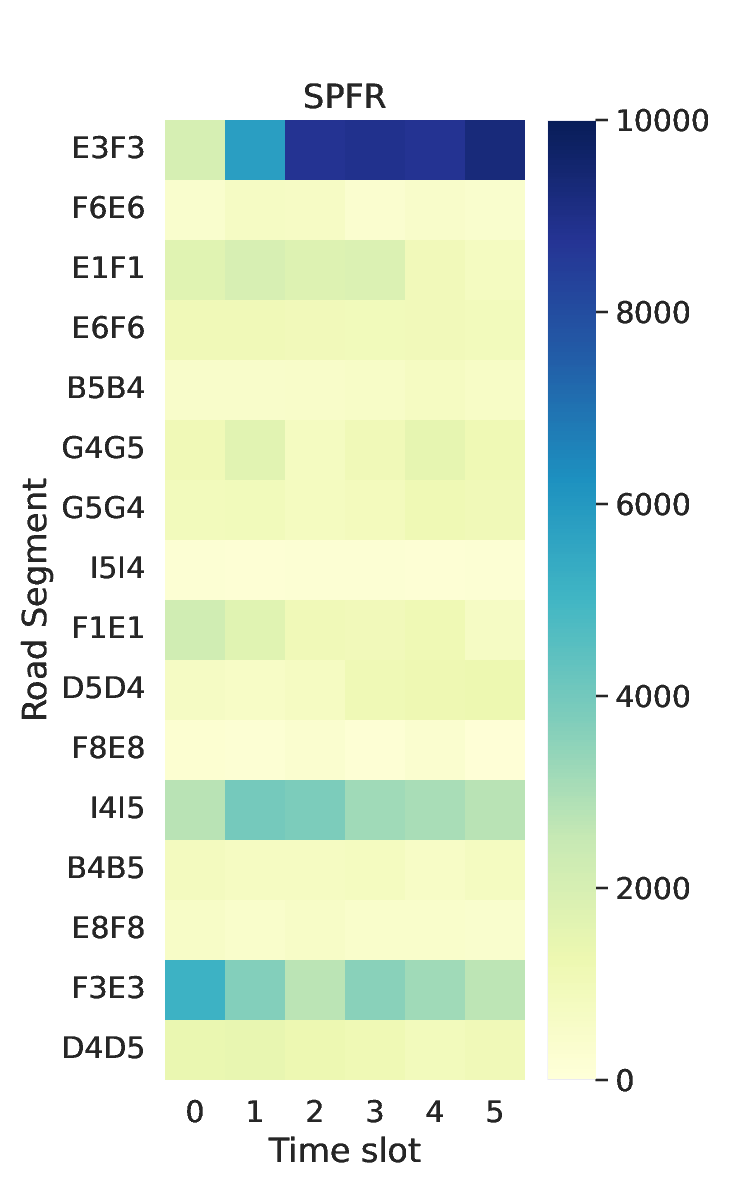}}
        \subfigure[Q-Routing]{
    \includegraphics[width=0.15\linewidth]{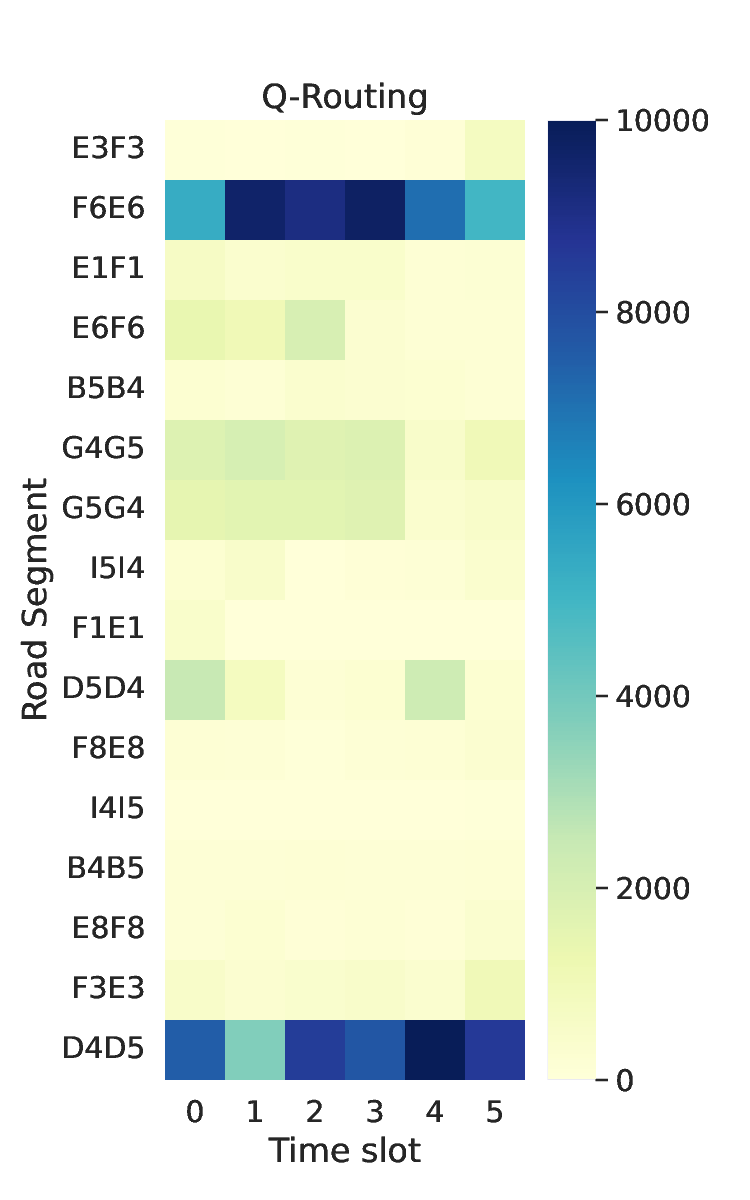}}
        \subfigure[AN]{
    \includegraphics[width=0.15\linewidth]{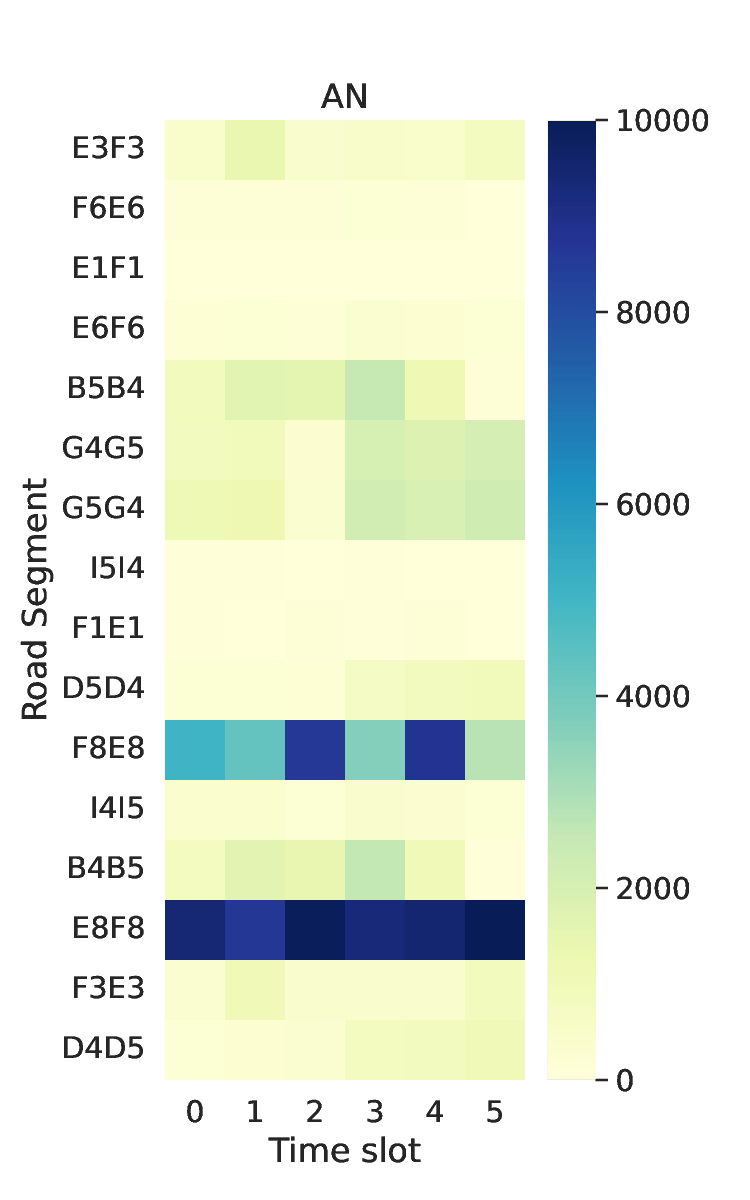}}
    \subfigure[asyn-MARL]{
    \includegraphics[width=0.15\linewidth]{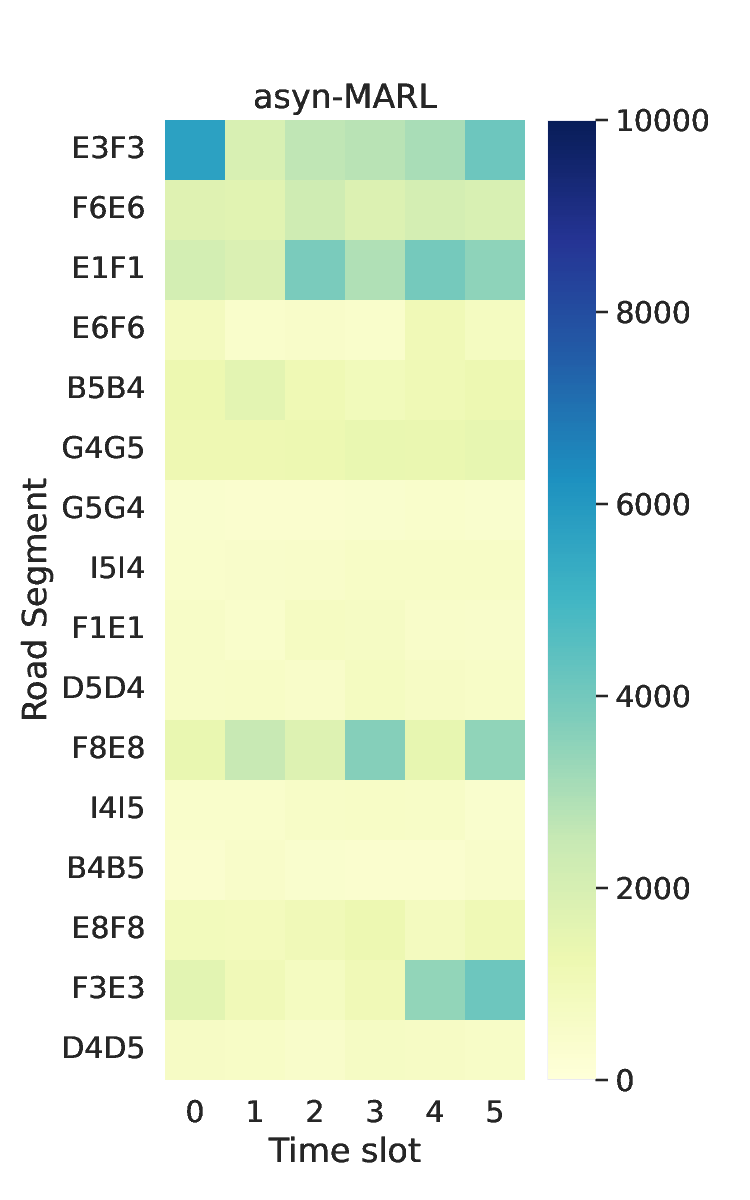}}\vspace{-2ex}
\caption{Distribution of vehicle volume on cutting edges with different routing methods {on synthetic data set}. }
\label{fig:load_balance}
\end{figure*}

In this section, we evaluate the load balance of all algorithms in terms of the distribution of traffic volumes (the number of vehicles) across the entire cutting edges. As shown in Figure \ref{fig:load_balance}, 
the $x$-axis and $y$-axis indicate the time slot during a span of the simulation with 5 minutes and the ID number of 16 cutting edges, which are marked in Figure \ref{fig:road_network}(a).
and the color darkness instead means the traffic volumes on the associated cutting edge within the time slot. As shown in this figure, our asyn-MARL demonstrates a rather even distribution of traffic volumes on the 16 cutting edges across the entire time slots. Instead. the SP and Random approaches tend to plan too many vehicles towards two cutting edges (E3F3 and F3E3), resulting in traffic congestion. Instead, SPFR could re-route some traffics towards other cutting edges such as I4I5, and lead to shorter travel time. In addition, though both Q-Routing and AN can re-route some traffics from the two cutting edges (E3F3 and F3E3) to alternative edges, such re-routed traffics are unfortunately re-directed to the same cutting edges, i.e., D4D5 in  Figure \ref{fig:ablation}(d) and  F8E8 in Figure \ref{fig:ablation}(e), again suffering from traffic congestion. In contrast, due to the proposed actor and critic networks, asyn-MARL has the chance to choose diverse cutting edges even for those vehicles with close destinations, and thereby mitigates traffic congestion for better load balancing.

\subsubsection{Impact of Graph Division}
In this section, we study the effects of graph division on the synthetic data set. By varying the number of regions from 3 to 6, we compare the effect of two graph division algorithms. (1) Random Approach: we randomly select $M$ nodes as the initial seed nodes of the $M$ regions, then iteratively add a neighbour node to each region until all the nodes are assigned to the associated regions. (2) We use the METIS \cite{karypis1997metis} algorithm to divide the road graph. This algorithm aims to create sub-graphs while minimizing the number of cutting edges and ensuring sub-graphs with similar size. Figure \ref{fig:road_network} colorizes the divided sub-graphs (regions) and we comfortably recognize that such division is roughly consistent with the overall graph topology.

\begin{figure}[ht]
\centering
    \subfigure[Throughput]{
    \includegraphics[width=0.47\linewidth]{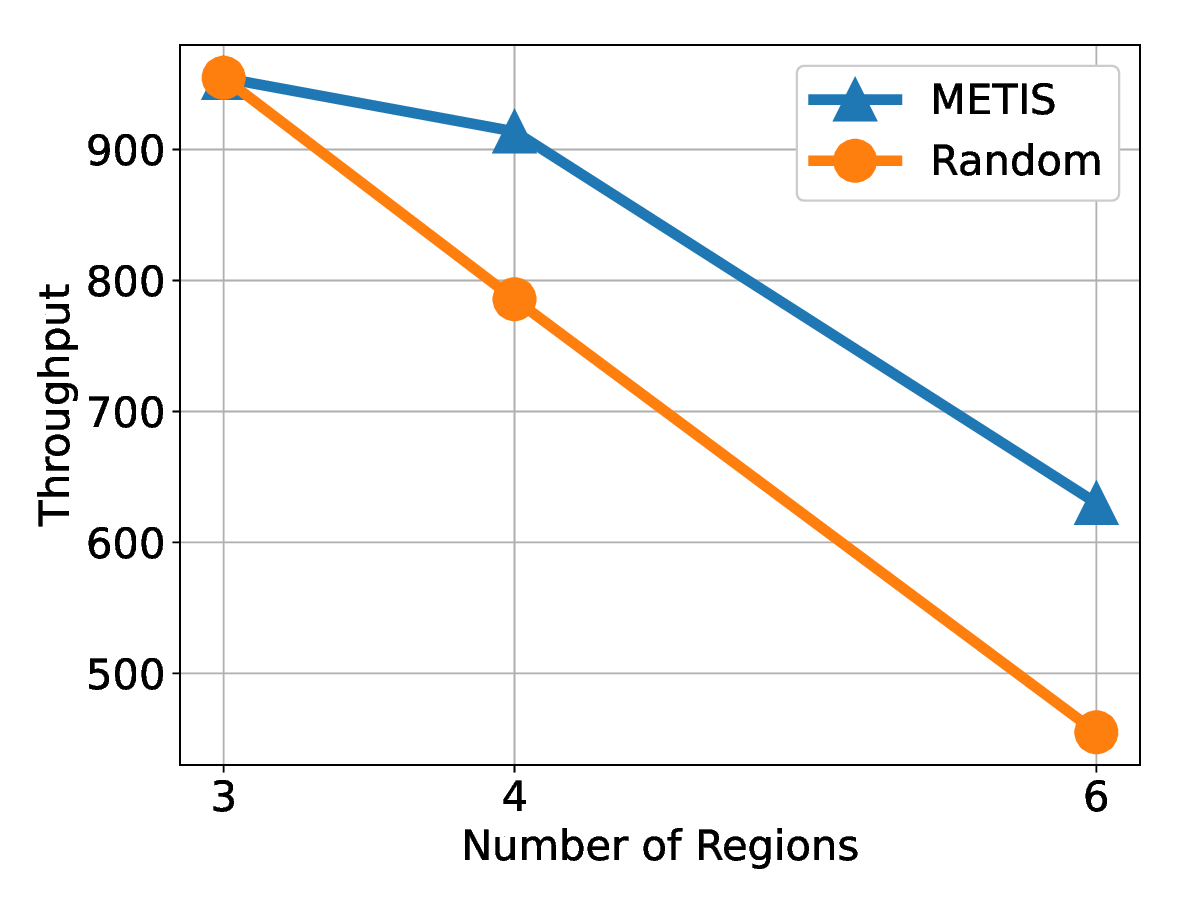}}
    \subfigure[Average Travel Time (second)]{
    \includegraphics[width=0.47\linewidth]{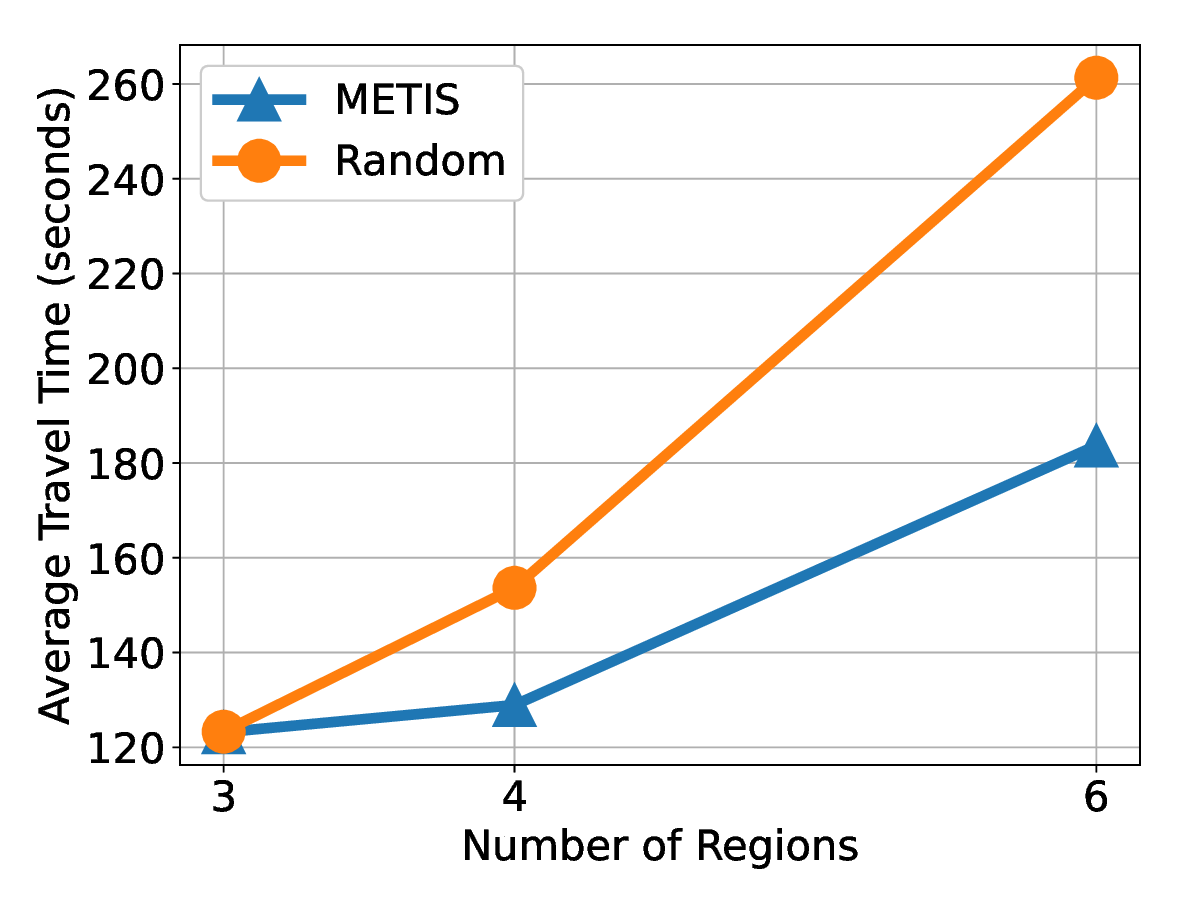}}\vspace{-2ex}
\caption{Effect of graph division approaches (Random and METIS)}
\label{fig:region}
\end{figure}

Figure \ref{fig:region} gives the performance under the two graph division approaches. We can observe that random division yields worse performance. That is, the random division generates more cutting edges and higher action space of region agents than METIS. Thus, it makes sense that the asyn-MARL model trained  on such divided sub-graphs suffers from worse result. Moreover, when the number of regions grows, the performance of asyn-MARL decreases, due to the following reason. Given a smaller number of regions, we have make the implicit assumption that a region agent has the higher capacity to observe the details of the road segments within in a larger region. Otherwise, when the number of subgraphs is greater, region agents can observe the details of a smaller region and yet have to roughly estimate the majority of the entire graph, leading to performance degrade. 

\subsubsection{Impact of Road Capacity Limit and Traffic Volume} \label{sec:sensitivity}
\begin{figure}[htbp]
\centering
    \subfigure[]{
    \includegraphics[width=4.0cm]{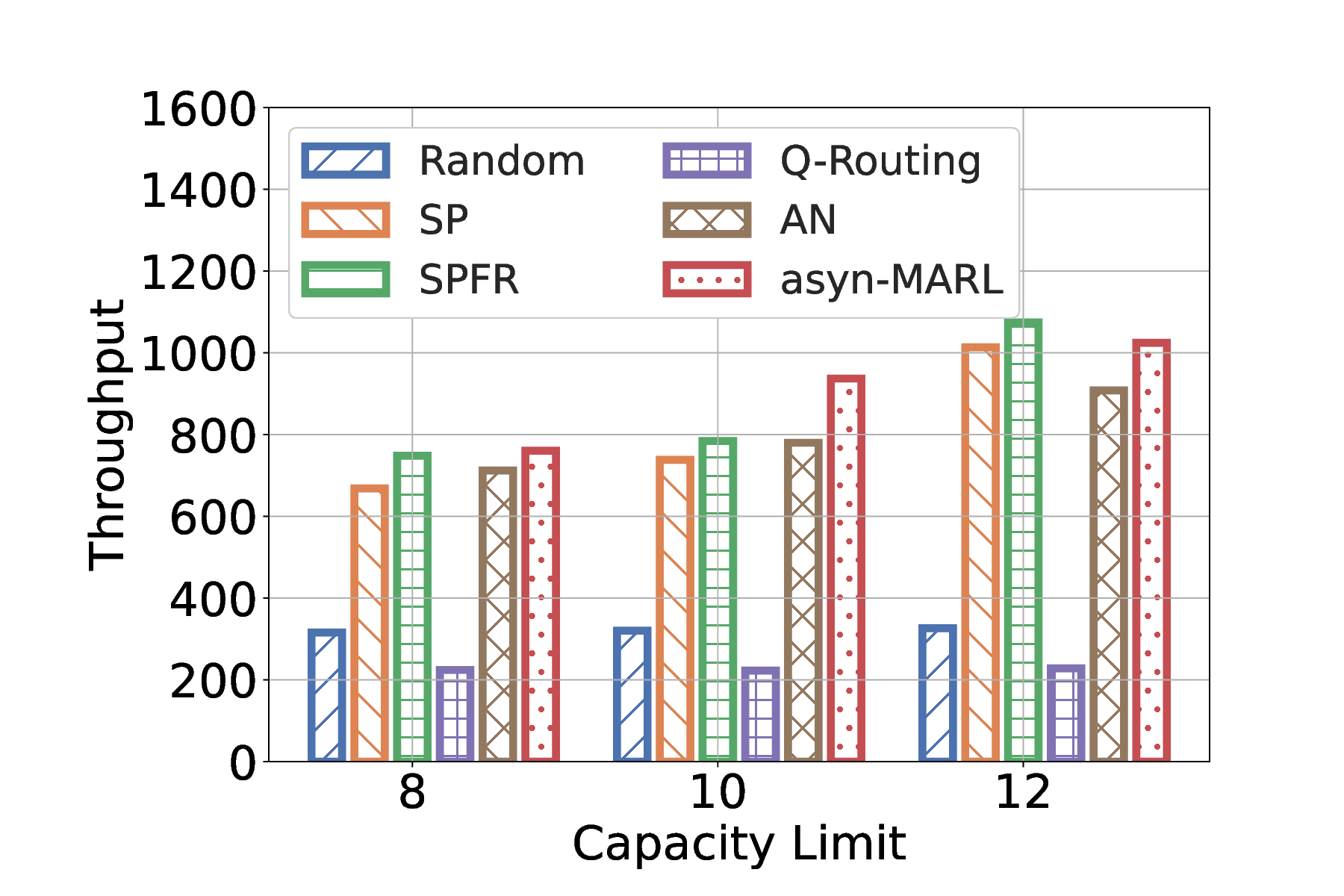}
    }
    \subfigure[]{
    \includegraphics[width=4.0cm]{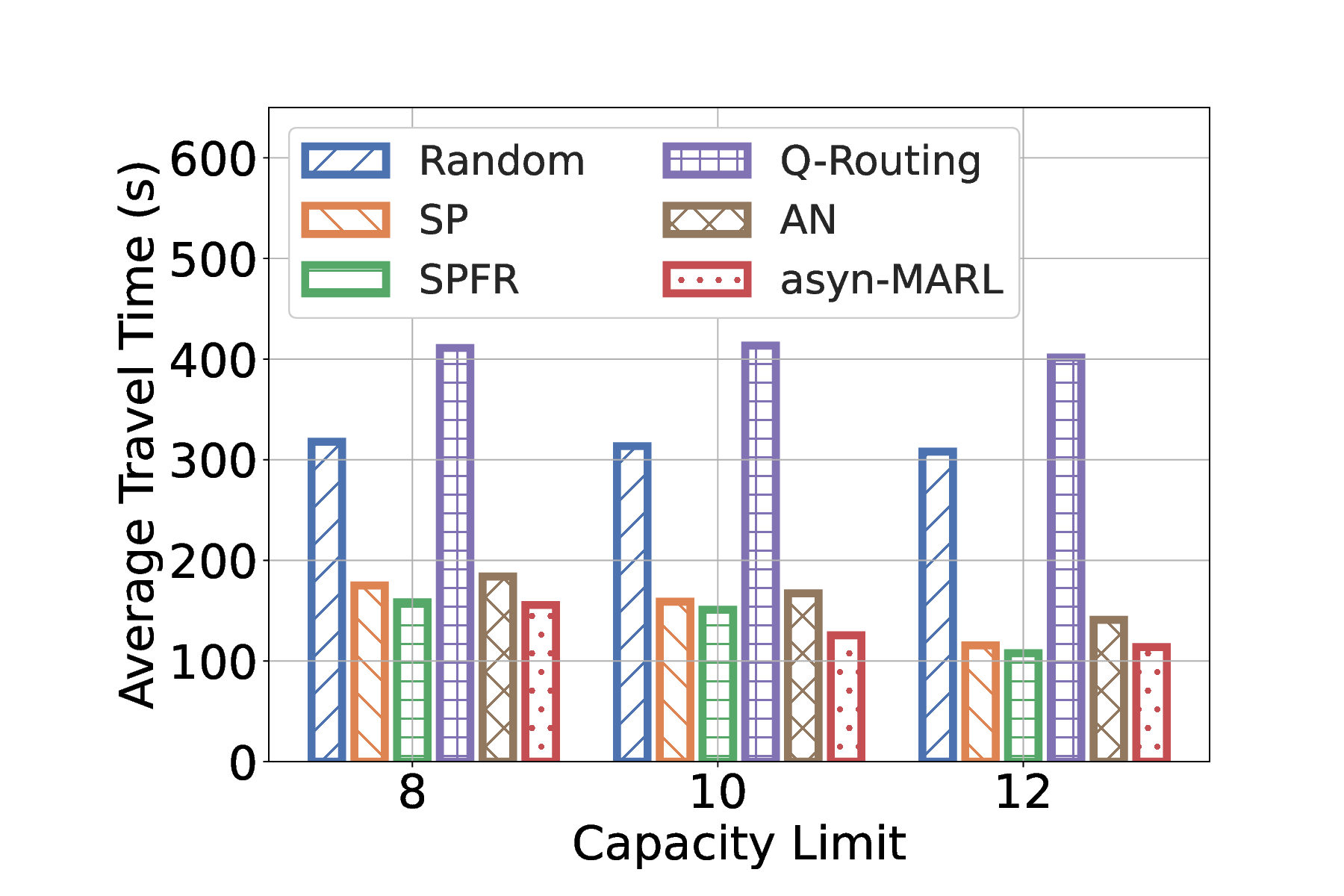}
    }
    \subfigure[]{
    \includegraphics[width=4.0cm]{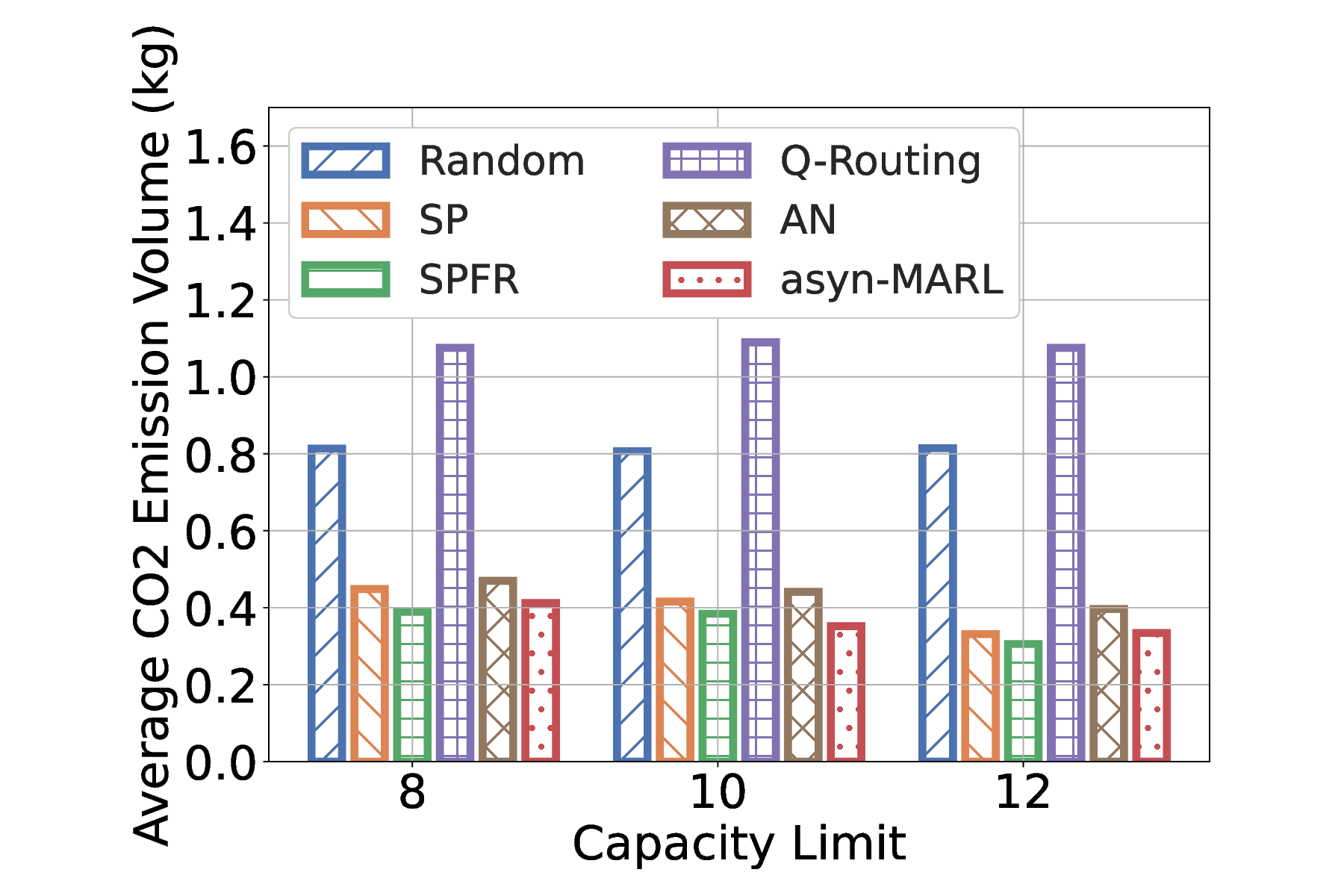}
    }
    \subfigure[]{
    \includegraphics[width=4.0cm]{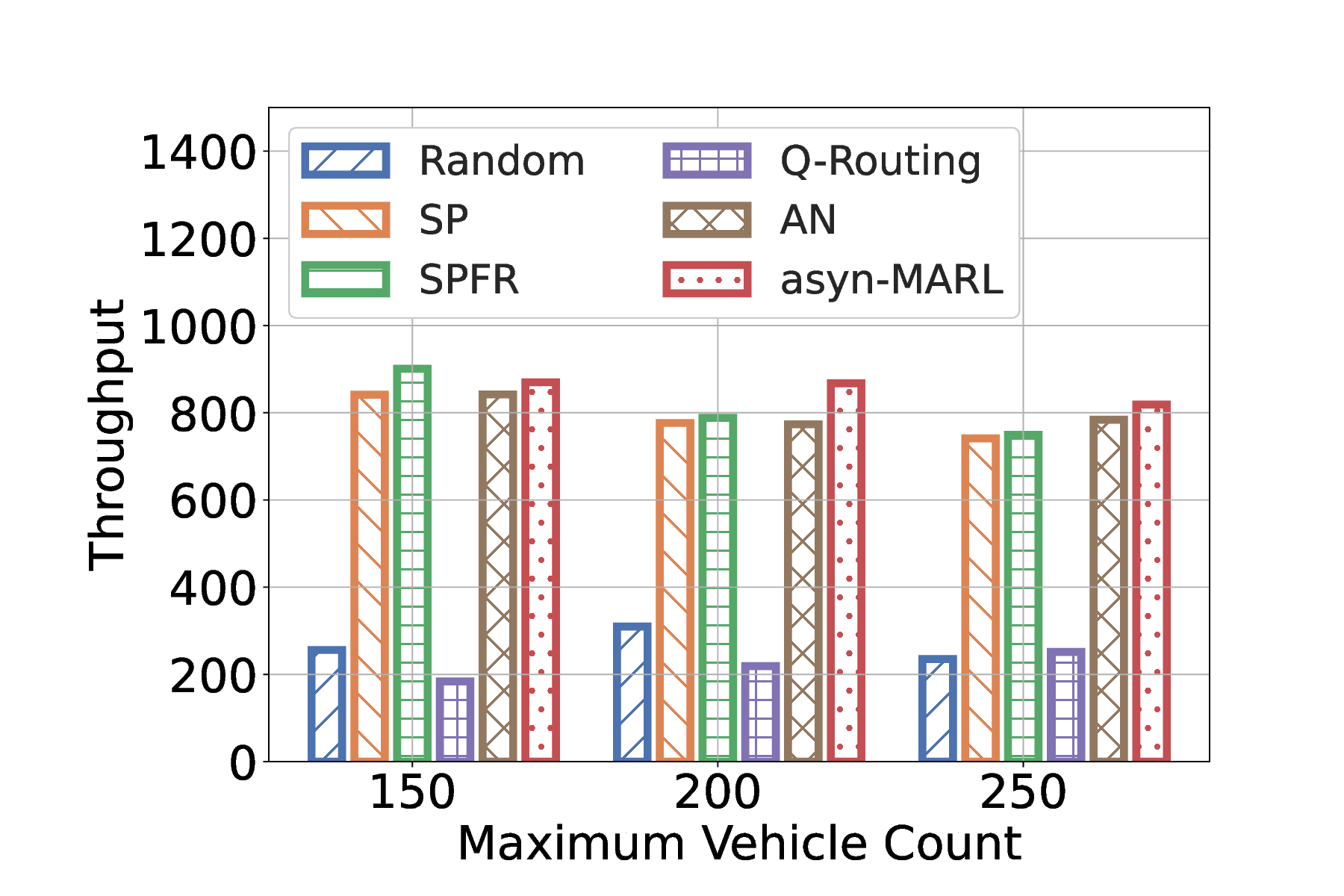}
    }
    \subfigure[]{
    \includegraphics[width=4.0cm]{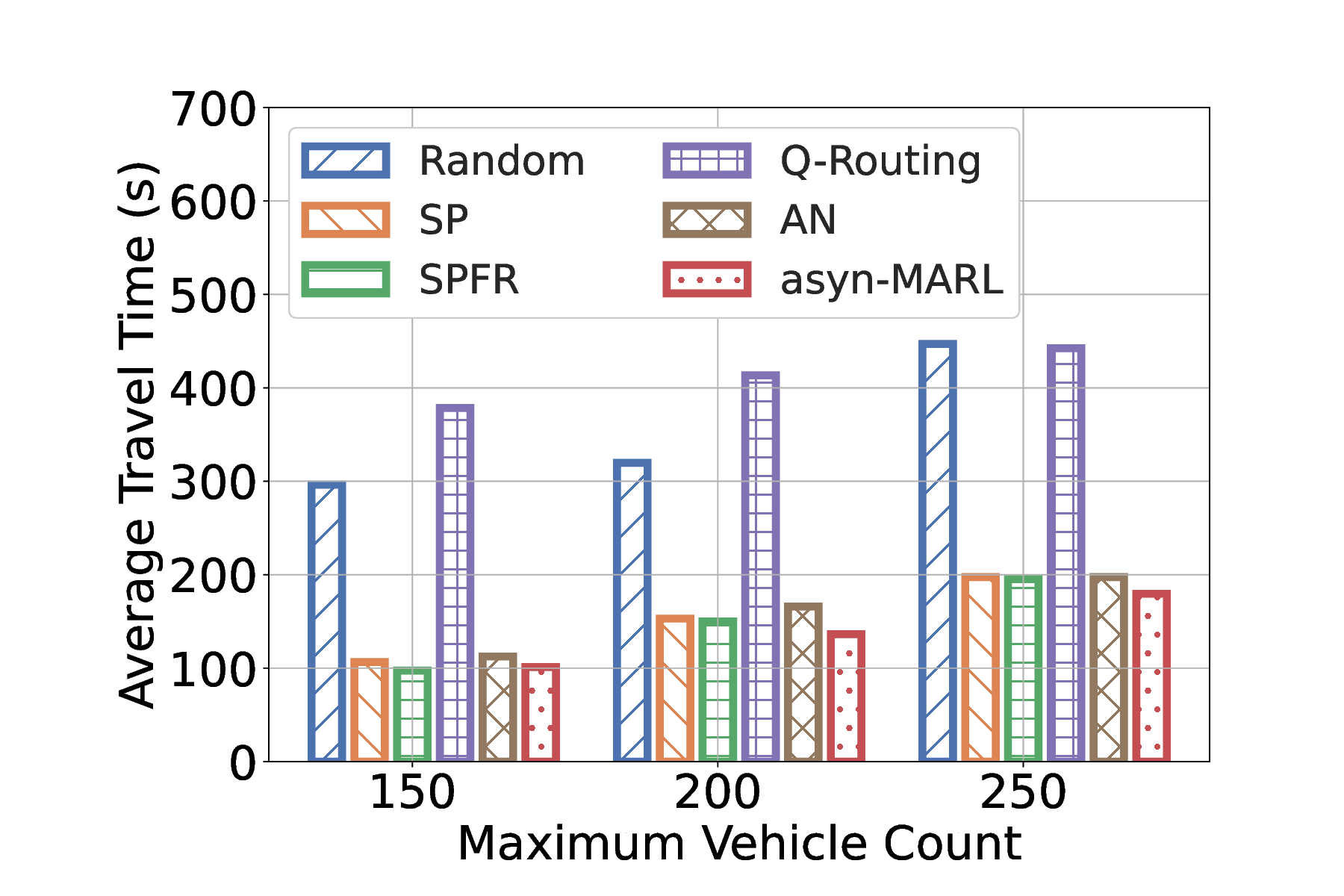}
    }
    \subfigure[]{
    \includegraphics[width=4.0cm]{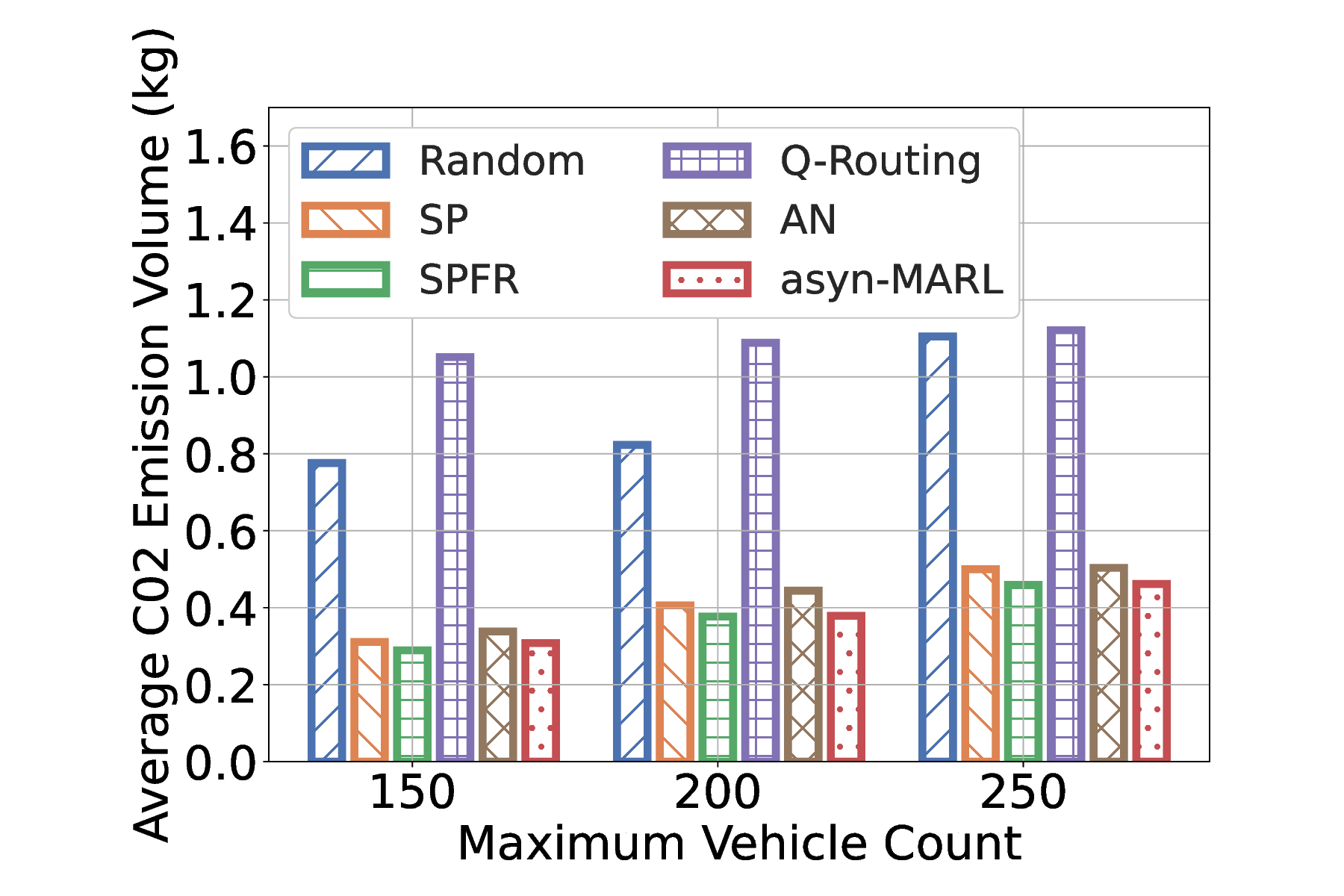}
    }\vspace{-2ex}
\caption{Impact of Road Capacity (a, b, c) and Traffic Volume (d, e, f).}
\label{fig:sensitivity}
\end{figure}

We finally evaluate the performance of our work under different traffic settings. Figure \ref{fig:sensitivity} (a-c) gives the performance of the six methods by varying the road capacity limits $c\in{\{8, 10, 12\}}$. We can find that when the capacity limit is smaller, i.e., the road segments are more likely to be congested, our work asyn-MARL consistently outperforms the competitors. Note that, for a small capacity limit such as 8, the perforance gap of our work and the competitor 
is significant. Such result indicates that our work asyn-MARL prefers to work well in the setting with limited resource.

Figure \ref{fig:sensitivity} (d-f) studies the performance of asyn-MARL when the vehicle count in the synthetic road network varies from 150 to 250. We can find that, when the vehicle count is smaller than 200, the performance of SP, SPFR and asyn-MARL is relatively close, because the road segments are less likely to be congested. As the vehicle count grows, the competition becomes more severe, and the average travel time of all approaches grows. In particular, in such a setting with more resource competition (e.g., caused by more vehiles on the road network), our work asyn-MARL outperforms others. The results again indicate that our work asyn-MARL works well in the setting. 

\section{Conclusion}\label{sec6:conclude}
In this paper, we propose a cooperative shortest path planning framework in the asynchronous MSD-SPP setting. To deal with the ineffectiveness and inefficiency issue of existing works, we propose the two-stage routing plans (inter-region and intra-region plans), and formulate the inter-region planning as a decentralized partial observable Markov decision process (Dec-POMDP). The proposed framework asyn-MARL is an actor-critic-based algorithm consisting of a centralized critic improved by asynchronous trajectory collection to address the non-stationary issue of MARL training, a novel actor to cooperate the vehicles with close destinations, and a reachability graph to avoid infinite loops. On both synthetic and real datasets, our evaluation demonstrates that our work asyn-MARL outperforms both the classic shortest path computation algorithms and the recent MARL-based baselines.

\bibliography{ref}

\bibliographystyle{abbrv}

\begin{IEEEbiography}
[
{
\includegraphics[width=1in,height=1.25in,clip,keepaspectratio]{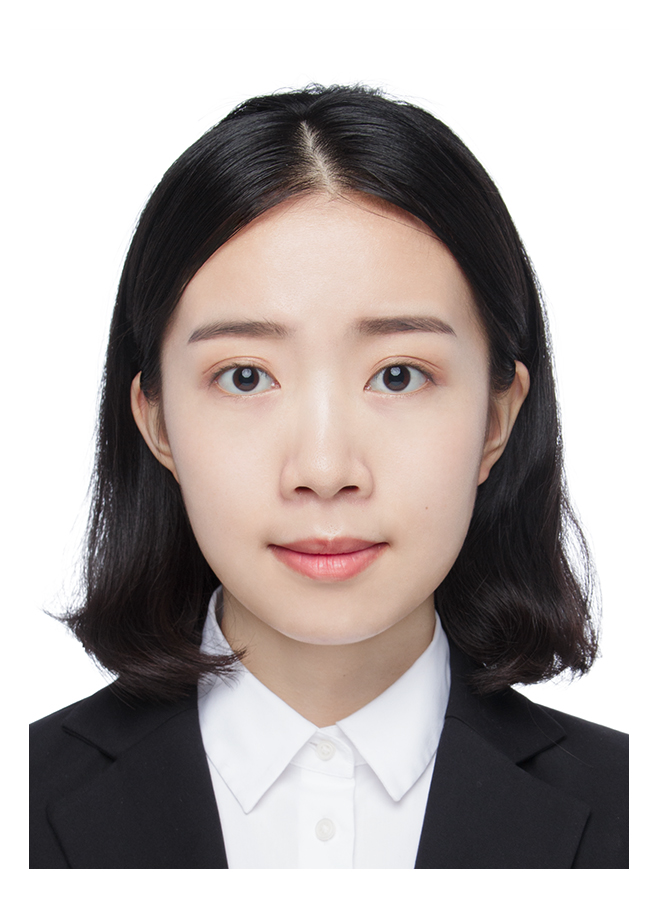}
}]
{Jiaming Yin}
received the B.E. degree from Tongji University, Shanghai, China, in 2019, where she is currently pursuing the Ph.D. degree with the School of Software Engineering. Her research interests include time series, reinforcement learning and mobile computing.
\end{IEEEbiography}

\begin{IEEEbiography}
[
{
\includegraphics[width=1in,height=1.25in,clip,keepaspectratio]{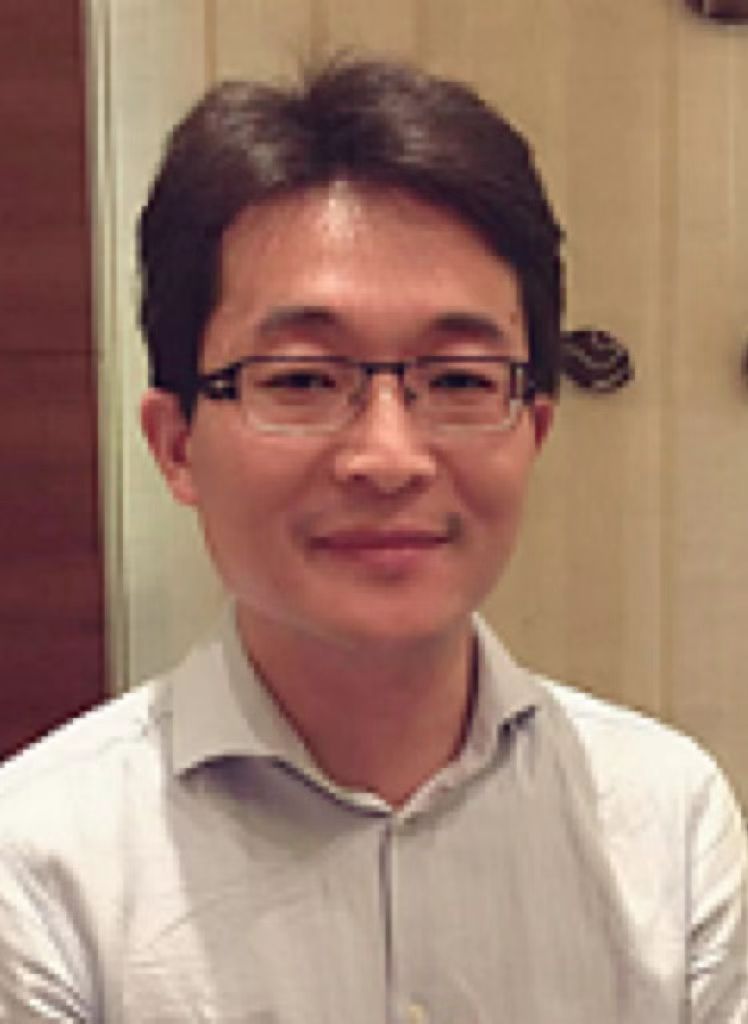}
}
]
{Weixiong Rao}
(Member, IEEE) received the PhD degree from The Chinese University of Hong Kong, Hong Kong, in 2009. After that, he worked for Hong Kong University of Science and Technology(2010), University of Helsinki (2011-2012), and University of Cambridge Computer Laboratory Systems Research Group (2013) as postdoctor researchers. He is currently a full professor with School of Software Engineering, Tongji University, China (since 2014). His research interests include mobile computing and spatiotemporal data science, and is a member of the CCF and ACM.
\end{IEEEbiography}

\begin{IEEEbiography}
[
{
\includegraphics[width=1in,height=1.25in,clip,keepaspectratio]{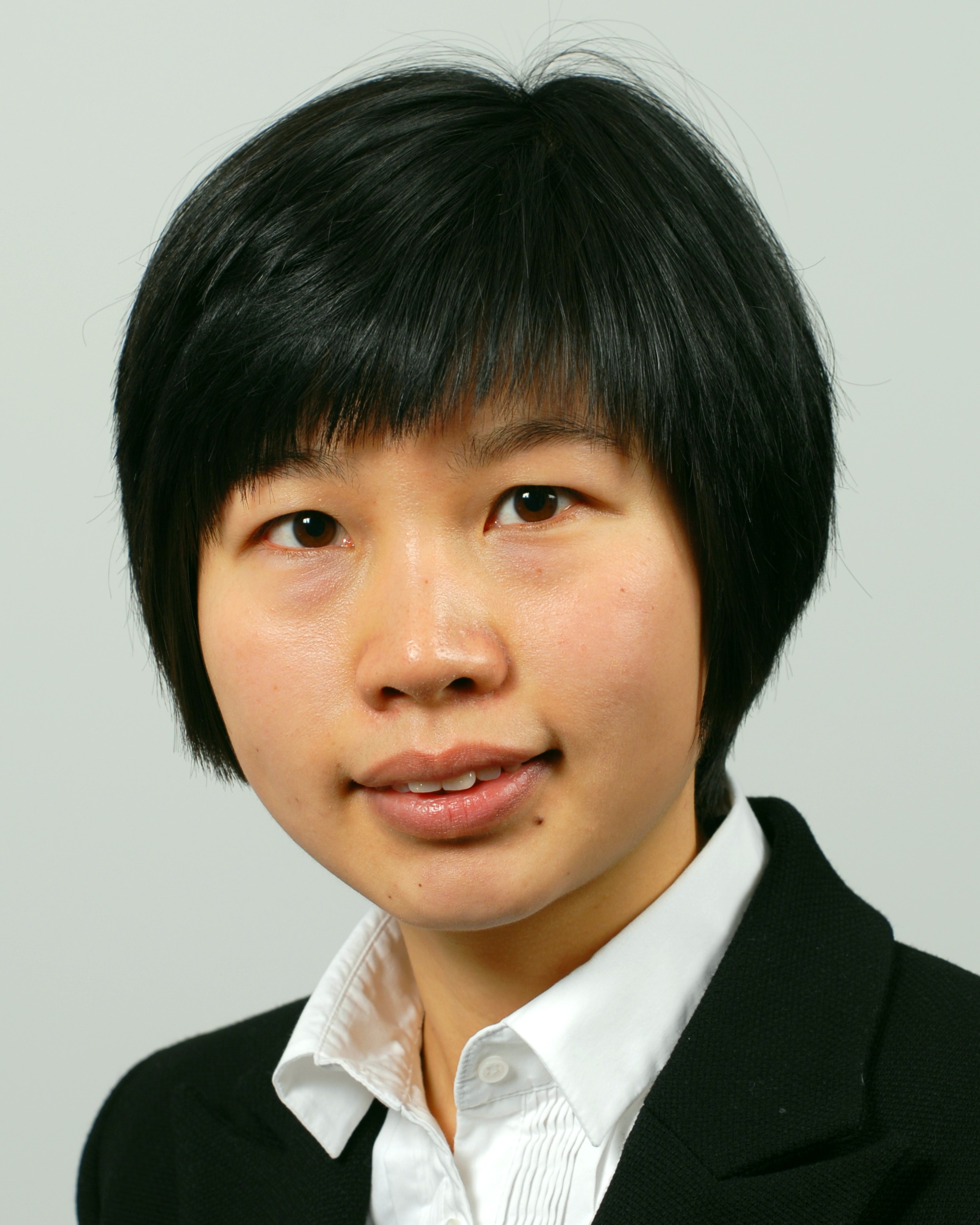}
}
]
{Yu Xiao}
(Member, IEEE) Yu Xiao received the doctoral degree in computer science from Aalto University, in 2012. She is currently an associate professor with the Department of Communications and Networking, Aalto University. Her current research interests include edge computing, wearable sensing and extended reality. She is a member of the IEEE.
\end{IEEEbiography}

\begin{IEEEbiography}
[
{
\includegraphics[width=1in,height=1.25in,clip,keepaspectratio]{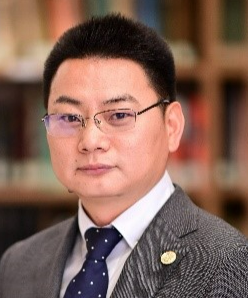}
}
]
{Keshuang Tang}
received the Ph.D. degree in transportation engineering from Nagoya University in 2008. He was a Post-Doctoral Research Fellow with The University of Tokyo. He was a Project Assistant Professor with Tohoku University. He is currently a Professor with the Department of Transportation Information and Control Engineering, Tongji University, China. His main research interests include driver behavior, signal control, and intelligent transportation systems.
\end{IEEEbiography}

\end{document}